%% file: main.tex
\documentclass[11pt]{scaleai-paper}

\usepackage{amsmath}
\usepackage{amsfonts}
\usepackage{amssymb}
\usepackage{amsthm}
\usepackage{booktabs}
\usepackage{tabularx}
\usepackage{tabulary}
\usepackage{multirow}
\usepackage{subcaption}
\usepackage{amsmath,amssymb,amsthm}
\usepackage{float}
\usepackage[square,numbers,sort&compress]{natbib}
\usepackage{xspace}
\usepackage{url}
\usepackage{fancyvrb}

\theoremstyle{remark}

\newcolumntype{L}{>{\RaggedRight\arraybackslash}p}
\usepackage{graphicx}
\usepackage{pgfplots}
\usepackage{ragged2e}
\usepackage{xcolor}

\pgfplotsset{compat=1.18}
\usetikzlibrary{arrows.meta,positioning}
\usepackage[colorlinks=true,linkcolor=scaleLink,citecolor=scaleLink,urlcolor=scaleLink]{hyperref}
\usepackage[capitalise,nameinlink]{cleveref}
\input{math_commands.tex}

\newcolumntype{Y}{>{\RaggedRight\arraybackslash}X}
\let\svthefootnote\thefootnote
\newcommand\freefootnote[1]{%
  \let\thefootnote\relax%
  \footnotetext{#1}%
  \let\thefootnote\svthefootnote%
}

\providecommand{\yes}{\textcolor{scaleForest}{\checkmark}}
\providecommand{\pmark}{\textcolor{scaleGold}{\ensuremath{\sim}}}
\providecommand{\nmark}{\textcolor{scaleMediumGray}{\ensuremath{\times}}}

\newcolumntype{Z}[1]{>{\RaggedRight\hyphenpenalty=10000\arraybackslash}p{#1}}

\DefineVerbatimEnvironment{efecode}{Verbatim}{%
  fontsize=\small, frame=single, framerule=0.4pt,
  rulecolor=\color{scaleMediumGray}, framesep=6pt,
  formatcom=\color{scaleInk}}

\contact{\texttt{veronica.chatrath@scale.com, yuan.xue@scale.com} \quad | \quad \url{https://scale.com/research}}

\title{\EFE{} or Not: \efename{}}

\author[1]{Veronica Chatrath\textsuperscript{*}}
\author[1]{Bryan Zhu\textsuperscript{*}}
\author[1]{Jingxuan Fan\textsuperscript{*}}
\author[1]{George Pu}
\author[1]{Soham Dinesh Tiwari}
\author[1]{Soham Dan}
\author[1]{Ryan Young}
\author[1]{Yuan (Christy) Li}
\author[1]{Yuang Yao}
\author[1]{Apaar Shanker}
\author[1]{Minglai Yang}
\author[1]{Daniel Yue Zhang}
\author[1]{Yunzhong He}
\author[1]{Ying Liu}
\author[1,2]{Chenguang Wang}
\author[1,3]{Zhijun Yin}
\author[1]{Yuan (Emily) Xue}

\affil[1]{Scale AI}
\affil[2]{University of California, Santa Cruz}
\affil[3]{Vanderbilt University Medical Center}
\affil[*]{Co-first authors.}

\usepackage[most]{tcolorbox}
\usepackage{xcolor}

\definecolor{todored}{RGB}{185,45,45}
\definecolor{todoorange}{RGB}{205,120,20}
\definecolor{todoblue}{RGB}{45,95,170}
\definecolor{todogreen}{RGB}{45,135,75}
\definecolor{todogray}{RGB}{90,90,90}

\newtcolorbox{efetodo}[2][]{
    enhanced,
    breakable,
    colback=gray!4,
    colframe=todored,
    coltitle=white,
    colbacktitle=todored,
    fonttitle=\bfseries,
    title={#2},
    boxrule=0.8pt,
    arc=2pt,
    left=6pt,
    right=6pt,
    top=5pt,
    bottom=5pt,
    #1
}

\begin{document}


\maketitle

\input{sections/abstract}
\input{sections/intro}
\input{sections/overview}
\input{sections/metrics}
\input{sections/standard}
\input{sections/results}
\input{sections/related}
\input{sections/limits}
\input{sections/conclusion}

\clearpage

\bibliographystyle{unsrtnat}
\bibliography{references}

\clearpage
\appendix
\input{sections/appendix}

\end{document}

%% file: math_commands.tex
\newcommand{\efename}{\emph{Reliable Enterprise Agent Deployment}}
\newcommand{\EFE}{READY}

%% file: sections/abstract.tex
\begin{abstract}
An AI agent can perform well on benchmarks and still be unsuitable for deployment. Existing AI-agent benchmarks primarily measure whether an agent can complete realistic professional work, whereas enterprise deployment requires asking a different question: whether an agent can meet a required reliability level, under an acceptable level of human oversight, and at a tolerable cost. We introduce \efename{} (\EFE{}{}), an evaluation framework for qualifying AI agents for deployment on concrete enterprise workflows. \EFE{} preserves each workflow's own definition of successful execution while applying a common deployment-qualification procedure. Given an agent, a workflow, and a class of candidate oversight policies, \EFE{} measures the reliability and operating cost of the human-AI system, selects the minimum-cost policy that satisfies a specified reliability target, and statistically qualifies the selected policy on held-out cases.  The \emph{deployment profile} characterizes the operating point supported by the evidence, including reliability, human-oversight burden, and cost. \EFE{} is implemented as an open testbed that decouples workflow specification, execution, evaluation, and deployment qualification, and runs on existing agent-evaluation infrastructure. In an end-to-end clinical-audit case study spanning 16 agent systems and 750 cases, \EFE{} reveals deployment differences hidden by autonomous benchmark performance: two systems separated by only 0.3 percentage points in autonomous accuracy (72.8\% vs.\ 72.5\%) require 39.2\% versus 29.6\% human review, respectively, to qualify at the same 76\% reliability target under the evaluated oversight policy. Systems with nearly identical autonomous performance can support substantially different reliability--oversight tradeoffs. Accordingly, \EFE{} shifts enterprise agent evaluation from asking only \emph{how well can the agent perform the work?} to asking \emph{under what conditions, and at what cost, can it be reliably deployed?} By making those conditions explicit and statistically testable, \EFE{} provides a practical basis for comparing agent systems, setting oversight requirements, and making evidence-based deployment decisions. 
\end{abstract}


%% file: sections/intro.tex
\section{Introduction}
\label{sec:introduction}
Enterprises are starting to hand real professional work to AI agents that reason over domain
knowledge, use tools, and carry out multi-step tasks. A growing number of benchmarks evaluate
these capabilities on professional workflows~\cite{agentslastexam,gdpval,xbench}, scoring
whether an agent can complete the task and how good the result is. 
These benchmarks mainly answer an important question about \emph{capability}. However, whether to deploy an agent is a different question. An organization rarely needs an agent to work unattended. Rather, it deploys a system in which some work may be handled autonomously while other work is reviewed, corrected, approved, or taken over by humans. A more relevant question is not how often the agent succeeds on its own, but whether the human--AI system around it can achieve the reliability a particular workflow requires, how much human oversight that takes, and what the operating policy costs. An agent that is right on 80\% of cases is not, by that fact alone, ready or unready to deploy. What matters is whether the wrong 20\% can be identified and sent to a human, and how much that review adds to the cost of running the system.

We introduce \efename{} (\EFE), a framework for turning workflow-level evaluation evidence into a deployment qualification. \EFE{} takes as input an agent, an enterprise workflow, a set of
representative cases, and a class of candidate oversight policies. Each workflow retains its own definition of successful execution, including outcome correctness, process requirements, evidence grounding, policy adherence, and other requirements appropriate to that form of work. \EFE{} then evaluates the human--AI system under candidate oversight policies and asks which policy can satisfy a specified reliability target at minimum operating cost. 

\begin{quote}
\centering
\emph{Professional-work benchmarks ask how well an agent can do the work autonomously.\\
\EFE{} asks under what conditions, and at what cost, an organization can reliably deploy
it.}
\end{quote}

At its core, \EFE{} casts deployment qualification as a constrained optimization problem over a class of oversight policies. For each workflow, the evaluator maps a multidimensional set of deployment-relevant criteria into a measure of successful execution, while the oversight policy specifies the points at which human intervention may occur. Agent executions on representative cases provide the empirical evidence for estimating the reliability and operating cost induced by each candidate policy. \EFE{} then searches the policy class for the lowest-cost operating policy that satisfies a specified reliability target and any additional deployment constraints. The selected policy is frozen and statistically qualified on held-out cases not used during policy selection. The resulting \emph{deployment profile} characterizes the reliability, oversight burden, cost, and workflow-specific risk of the qualified human--AI configuration.

We use a retrospective clinical-audit workflow~\cite{clinicare} as a running case study to illustrate \EFE{} end to end. In this setting, an agent receives a longitudinal patient record and a clinical audit question, such as whether an ICU patient developed acute kidney injury within 48 hours of a CT scan. It must identify the relevant evidence in the record, apply the governing clinical standard, and return a verdict together with a stated confidence. A capability benchmark stops at the verdict and scores it. However, deployment requires an additional decision: which completed audits can be accepted automatically and which should be escalated to a clinician. That oversight policy, rather than raw verdict accuracy alone, determines the reliability of the human--AI system, the review burden it imposes, and its operating cost.

Across 16 agent systems and 750 cases, the deployment profiles reveal differences that autonomous benchmark performance alone does not. For example, two systems separated by only 0.3 percentage points in autonomous accuracy (GPT-5.4's 72.8\% versus Sonnet 5's 72.5\%) require 39.2\% versus 29.6\% human review, respectively, to qualify at the same 76\% reliability target under the evaluated oversight policy. More generally, the reliability--oversight frontier shows that systems with similar autonomous performance can support substantially different operating points. On a conventional leaderboard, the two systems appear nearly identical, yet their deployment costs differ substantially.

\paragraph{What \EFE{} adds.}
\EFE{} complements rather than replaces existing agent evaluations. Capability and work-product benchmarks~\cite{agentslastexam,gdpval,xbench} measure the quality of agent execution but do not determine an operating point for a deployed human--AI system.
The \emph{pass\textsuperscript{k}} metric of $\tau$-bench~\cite{taubench} measures consistency across repeated attempts but does not incorporate human oversight or operating cost. Cost-aware routing and cascades~\cite{frugalgpt2023,ong2025routellm} optimize quality--cost tradeoffs by routing among models, while selective prediction and learning-to-defer study when a model should abstain or pass a case to a human~\cite{kapoor2024agentsmatter}. \EFE{} brings
these ideas into a workflow-level deployment setting: it evaluates the reliability and cost of the complete human--AI system, optimizes an oversight policy against an explicit reliability requirement, and statistically qualifies the selected operating point on held-out evidence.
It reuses the evidence produced by existing workflow evaluations instead of displacing them.

\paragraph{An open, extensible testbed.}
To support heterogeneous enterprise workflows, \EFE{} separates workflow-specific definitions of successful work from a common deployment-qualification procedure. We implement \EFE{} as an open, extensible testbed built on existing agent-evaluation infrastructure~\cite{inspect_ai}. Workflows may differ in their task populations, execution environments, evaluation criteria, and forms of oversight while sharing the same qualification abstraction. We provide a seed workflow to demonstrate complete end-to-end use of this interface, and new workflows, cases, environments, and evaluators can be contributed while retaining their own standards of correct work.

Our contributions are as follows.

\begin{enumerate}

\item \textbf{A deployment-centered evaluation objective.} We formulate agent deployment as evaluation of a human--AI system under an oversight policy,
moving from the question of how well an agent performs autonomously to which policy can meet a workflow-specific reliability requirement and at what operating cost
(Section~\ref{sec:overview}).

\item \textbf{A method for optimizing and statistically qualifying oversight.}
\EFE{} selects a minimum-cost oversight policy subject to a reliability target and qualifies the frozen policy on held-out data. The deployment profile characterizes the reliability--oversight--cost operating point supported by the evidence, not autonomous capability alone (Section~\ref{sec:qualification}).

\item \textbf{An open evaluation interface and end-to-end empirical instantiation.}
We provide an extensible interface that separates workflow evaluation from deployment
qualification and instantiate it on a retrospective clinical-audit workflow~\cite{clinicare}
across 16 agent systems and 750 cases, demonstrating that similar autonomous performance can
correspond to substantially different reliability--oversight tradeoffs
(Sections~\ref{sec:platform}--\ref{sec:empirical}).

\end{enumerate}

%% file: sections/overview.tex
\section{Design Overview}
\label{sec:overview}

We now formalize the \EFE{} deployment-qualification problem, which asks whether an agent can be deployed reliably and economically on a concrete enterprise workflow. For a given workflow and agent, \EFE{} evaluates the human--AI system induced by an \emph{oversight policy}, $\pi$. The policy specifies when the agent may operate autonomously and when human intervention may be introduced. Depending on the workflow, intervention may occur after a completed result or during execution through review, approval, clarification, correction, retry, escalation, or takeover. At a high level, \EFE{} separates three phases:

\begin{center}
\textbf{Workflow Evaluation}
$\;\longrightarrow\;$
\textbf{Policy Optimization}
$\;\longrightarrow\;$
\textbf{Deployment Qualification}.
\end{center}

\begin{figure*}[t]
    \centering
    \includegraphics[width=\linewidth]{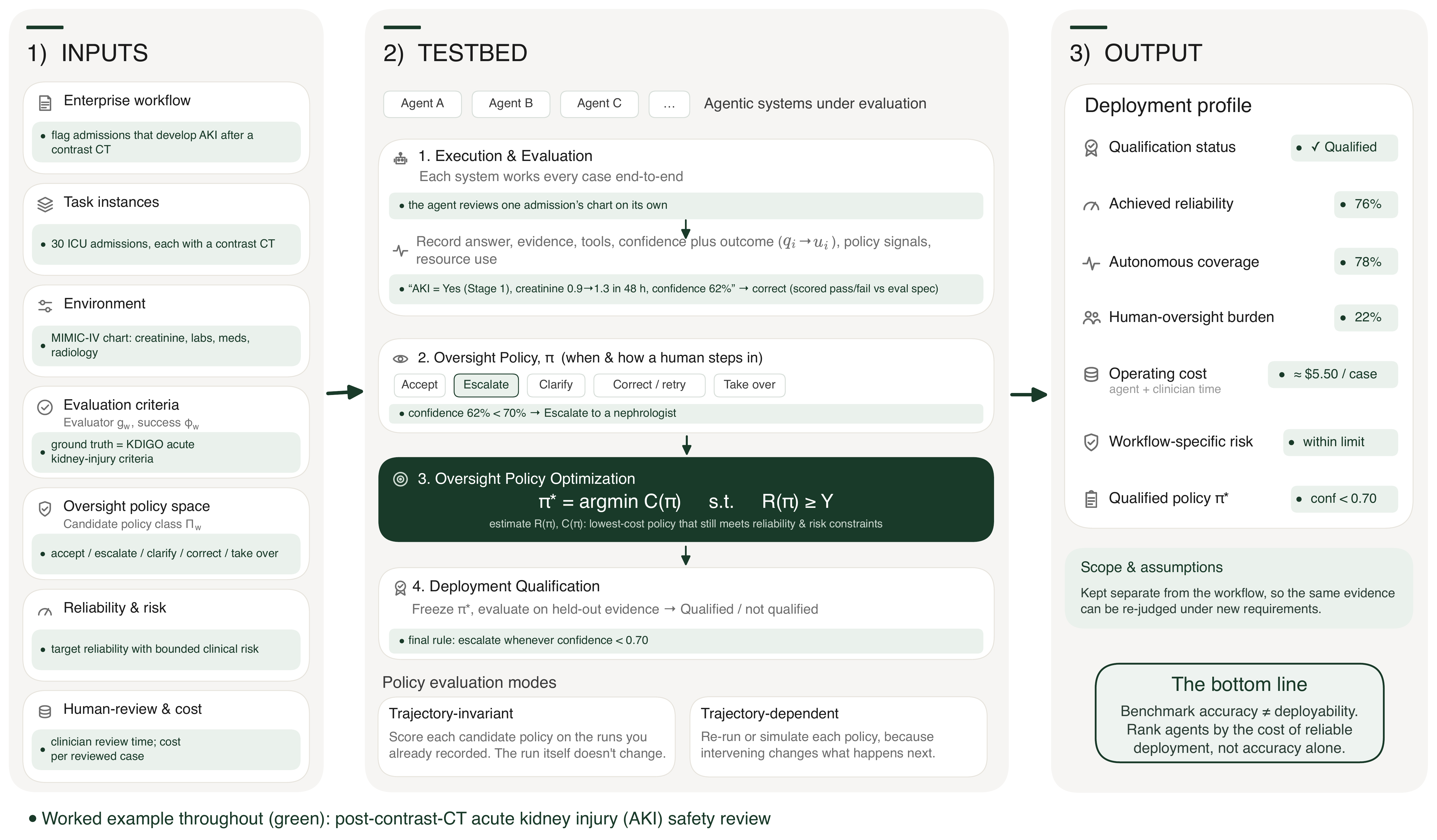}
\caption{
\efename{} (\EFE{}) takes as input a specification defining the workflow, representative task instances, execution environment, evaluation semantics, oversight policy class, and deployment requirements. Within the testbed, multiple agentic systems can be evaluated against the same specification. For each system, \EFE{} proceeds through three phases: (1) agent execution and workflow evaluation generate instance-level execution evidence; (2) oversight policy optimization selects the lowest-cost policy satisfying the specified reliability and risk constraints; and (3) statistical qualification freezes the selected policy and evaluates it on held-out cases. The output is a deployment profile for each evaluated system, characterizing the reliability, human-oversight burden, operating cost, and workflow-specific risk supported by the evidence under the stated deployment assumptions.
}
\label{fig:efe_overview}
\end{figure*}

Workflow evaluation defines and measures what constitutes successful execution for a particular form of professional work. Policy optimization uses this evidence to select, from a specified class of oversight policies, the lowest-cost policy that satisfies the deployment requirements. Deployment qualification then freezes the selected policy and statistically evaluates it on held-out evidence. The output of this process is a \emph{deployment profile} describing the reliability--oversight--cost operating point supported by the evidence under the stated deployment assumptions. Figure~\ref{fig:efe_overview} summarizes the end-to-end process.

\subsection{Evaluation Setting and Reliable Deployment Objective}
\label{sec:overview-setting}

A workflow, $w$, denotes a reusable type of enterprise work. Let $E_w$ denote the execution environment associated with workflow $w$, including the tools, data interfaces, systems, and interaction capabilities available during execution. A task instance, $x$, specifies one concrete case on which the workflow is performed. It contains the inputs made available to the agent, together with the reference material and ground truth required to evaluate that instance. Let $\mathcal{D}_w$ denote the population distribution over task instances of the workflow. An evaluation set, $S_w$, is a finite sample of instances,

\begin{equation}
S_w
=
\{x_i\}_{i=1}^{N_w},
\qquad
x_i \overset{\text{iid}}{\sim} \mathcal{D}_w .
\label{eq:workflow-dataset}
\end{equation}

Let $m$ denote the evaluated agent.
Executing $m$ on task instance, $x_i$, in environment, $E_w$, under oversight
policy, $\pi$, produces a trajectory:

\begin{equation}
Z_i^\pi
=
\operatorname{Execute}(m,x_i,E_w;\pi).
\label{eq:policy-trajectory}
\end{equation}

The trajectory contains the final work product and the observable evidence needed to evaluate the work, including, when relevant, retrieved evidence, tool interactions, intermediate artifacts, interventions, state changes, and resource use. If the agent, environment, or human oversight is stochastic, $Z_i^\pi$ and all derived quantities are random variables under the corresponding deployment process.

Each workflow defines a workflow-specific evaluator, $g_w$, that maps a trajectory to a vector of measurements:

\begin{equation}
\mathbf{q}_i^\pi
=
g_w\!\left(Z_i^\pi\right),
\label{eq:workflow-quality}
\end{equation}

where $\mathbf{q}_i^\pi$ may characterize both the final outcome and relevant aspects of the execution process, such as evidence use, adherence to required procedures, or compliance with applicable policies.
The workflow also defines a pre-specified
\emph{workflow-success predicate}, $\phi_w$:

\begin{equation}
u_i^\pi
=
\phi_w(\mathbf{q}_i^\pi)
\in
\{0,1\},
\label{eq:workflow-success}
\end{equation}

where $u_i^\pi = 1$ indicates that final outcome and any required process or policy conditions are satisfied on instance $i$. The vector $\mathbf{q}_i^\pi$ retains multidimensional diagnostic information about the execution, while $u_i^\pi$ is an instance-level execution outcome used to define deployment reliability.

This separation distinguishes measurement from qualification: $g_w$ determines what aspects of the work are measured, while $\phi_w$ specifies which combination of those measurements constitutes a successful execution. For professional workflows, the success predicate may require a conjunction of outcome correctness with selected process, grounding, or policy requirements, not final-answer correctness alone:

\begin{equation}
Z_i^\pi
\;\xrightarrow{\;g_w\;}\;
\underbrace{\mathbf{q}_i^\pi}_{\text{workflow measurements}}
\;\xrightarrow{\;\phi_w\;}\;
\underbrace{u_i^\pi}_{\text{instance-level success}}
\label{eq:evaluation-to-deployment}
\end{equation}

For agent $m$ on workflow $w$ under oversight policy $\pi$, we define deployment reliability as

\begin{equation}
R_{m,w}(\pi)
=
\mathbb{E}_{x \sim \mathcal{D}_w}\footnote{We simplify the notation here. Strictly speaking, the expectation is taken over both the workflow population $x\sim\mathcal{D}_w$ and any stochasticity in agent execution or human oversight.}
\left[
u^\pi(x)
\right].
\label{eq:efe-overview-reliability}
\end{equation}

Let $K_w(Z^\pi)$ denote the workflow-specific operating cost associated with
an execution under policy $\pi$, including agent execution cost and the cost of human oversight. The expected operating cost is

\begin{equation}
C_{m,w}(\pi)
=
\mathbb{E}_{x \sim \mathcal{D}_w}
\left[
K_w(Z^\pi)
\right].
\label{eq:efe-overview-cost}
\end{equation}

\EFE{} then casts oversight-policy selection as a constrained optimization
problem over a candidate policy class $\Pi_w$. Given a target reliability $Y$,
and optionally a workflow-specific risk tolerance $B_w$, define the feasible
set

\begin{equation}
\mathcal{F}_{m,w}(Y,B_w)
=
\left\{
\pi \in \Pi_w :
R_{m,w}(\pi)\geq Y,\;
\rho_w\!\left(L_w(Z^\pi)\right)\leq B_w
\right\},
\label{eq:efe-overview-feasible-set}
\end{equation}

where $L_w$ is a workflow-specific deployment loss and $\rho_w$ is a corresponding risk functional. The risk constraint is optional; when mean reliability is sufficient, it is inactive, equivalently $B_w=+\infty$.
For workflows in which the severity or concentration of failures matters,
$\rho_w$ may instead represent a chance constraint or a tail-risk measure such
as Conditional Value at Risk (CVaR).

If $\mathcal{F}_{m,w}(Y,B_w)=\varnothing$, no policy in the evaluated policy
class supports the requested deployment operating point. Otherwise, \EFE{}
selects the lowest-cost feasible policy:

\begin{equation}
\boxed{
\pi^{\star}_{m,w}(Y,B_w)
\in
\arg\min_{\pi\in\mathcal{F}_{m,w}(Y,B_w)}
C_{m,w}(\pi).
}
\label{eq:efe-overview-objective}
\end{equation}

Importantly, Equation~\ref{eq:efe-overview-objective} defines \emph{policy selection}, not statistical qualification. The policy is selected using development evidence, then frozen and evaluated on held-out qualification data to determine whether the requested reliability and any additional deployment constraints are statistically supported. Section~\ref{sec:qualification} develops the estimation of reliability, operating cost, risk, and statistical
qualification in detail.

\subsection{End-to-End \EFE{} Flow}
\label{sec:overview-flow}
Figure~\ref{fig:efe_overview} organizes \EFE{} around its inputs, three evaluation phases, and resulting deployment profiles. The input to \EFE{} is a specification defining the workflow, its task instances, associated execution environment, oversight policy class and evaluation criteria, including both the workflow-specific evaluator and the instance-level success criterion. Within the \EFE{} testbed, a set of agentic systems is evaluated against this specification. For each system, \EFE{} executes and evaluates the workflow, optimizes an oversight policy, and statistically qualifies the selected policy on held-out cases. The output is a deployment profile for each evaluated system.

\paragraph{Inputs: specification.}

\EFE{} begins from declarative specifications that define the deployment
question to be evaluated. The \emph{workflow specification} identifies the
workflow $w$, task population $\mathcal{D}_w$, sampled evaluation instances
$S_w$, and execution environment $E_w$. The \emph{evaluation specification}
defines the workflow evaluator $g_w$, the workflow-success predicate
$\phi_w$, and the observable signals or state available to oversight policies.
The \emph{deployment specification} defines the candidate policy class
$\Pi_w$, target reliability $Y$, human-oversight model, operating-cost model,
and any workflow-specific risk tolerance. Keeping these specifications
separate allows the same workflow evidence to be interpreted under different
deployment requirements.

\paragraph{Phase 1: agent execution and workflow evaluation.}

\EFE{} can evaluate multiple agent systems $m$ against the same deployment specification, producing comparable deployment profiles for each. For each evaluated system $m$ and task instance $x_i$, \EFE{} executes the agent in the
workflow environment $E_w$ and records the trajectory $Z_i$, including the
final work product, tool interactions, retrieved evidence, intermediate
artifacts, interventions, and resource use. The workflow evaluator maps this
trajectory into workflow-specific measurements,
$Z_i \xrightarrow{g_w} \mathbf{q}_i$, and the success predicate maps those
measurements into an instance-level success outcome,
$\mathbf{q}_i \xrightarrow{\phi_w} u_i$. The resulting evidence also includes
policy-observable signals and cost measurements needed to evaluate candidate
oversight policies.

\paragraph{Phase 2: oversight policy optimization.}
Using development evidence, \EFE{} estimates the reliability, operating cost, and any workflow-specific risk induced by candidate policies in $\Pi_w$. It then selects the lowest-cost policy satisfying the specified deployment constraints:
\[
\pi^\star
\in
\arg\min_{\pi \in \Pi_w} C(\pi)
\quad
\text{subject to}
\quad
R(\pi) \geq Y,
\]
together with any additional risk constraint. When oversight is
trajectory-invariant, such as terminal accept-or-escalate review, candidate
policies can be evaluated by replaying saved execution evidence. When
oversight is trajectory-dependent, intervention changes what happens next, 
candidate policies must be executed or simulated in the runtime.

\paragraph{Phase 3: deployment qualification.}
Policy optimization alone does not establish deployability. After the policy
$\pi^\star$ is selected, \EFE{} freezes it and evaluates the human--AI system on held-out qualification cases not used during policy
selection. Deployment qualification is granted only when the held-out evidence
statistically supports the specified reliability target and any additional
deployment constraints.

\paragraph{Output: deployment profiles.}

The output of \EFE{} is a set of \emph{deployment profiles}, one for each
evaluated agentic system. A deployment profile reports the qualified policy,
qualification status, achieved reliability and statistical confidence,
human-oversight burden, operating cost, workflow-specific risk, and the scope
and assumptions under which the profile holds. The deployment profile connects benchmark evidence to an operational decision: whether, and under what oversight policy, the agent can be deployed at the required level of reliability and cost.

Section~\ref{sec:qualification} develops the qualification methodology in
detail, while Section~\ref{sec:platform} describes the evaluation runtime
and open interfaces used to implement this architecture across enterprise
workflows.

\subsection{Running Example: Retrospective Clinical Audit}
\label{sec:overview-clinicare}

Throughout the paper, we use a retrospective clinical-audit workflow from
CliniCARE-Bench~\cite{clinicare} as a running example to instantiate the
\EFE{} framework. 

In CliniCARE-Bench, for each patient-specific case, an agent receives a concise clinical-audit
question and access to a longitudinal patient record. It must determine what
evidence is needed, retrieve and reconcile that evidence, apply the relevant
clinical or operational standard, and produce an auditable adjudication.

In \EFE{} notation, retrospective clinical audit defines the workflow $w$. A
task instance $x_i$ consists of a clinical-audit question paired with the
relevant patient record and, where applicable, a particular encounter or
clinical event. The environment $E_w$ provides patient-scoped access to
structured and free-text EHR data through clinically meaningful retrieval
tools, together with facilities for explicit computation and access to
governing policy documents.

Executing agent $m$ on case $x_i$ produces a complete evaluation run. The
final report contains one of four verdicts---\emph{Yes}, \emph{No},
\emph{Indeterminate: Lack of Data}, or \emph{Indeterminate: Medically
Ambiguous}---together with supporting patient- and policy-evidence citations
and a stated confidence $s_i$. The recorded trajectory $Z_i$ additionally
contains observable investigation evidence such as tool calls, retrieved
information, computations, and intermediate artifacts.

For example, one scenario asks whether an ICU patient developed acute kidney injury within 48 hours after a CT scan. The agent must establish baseline creatinine using a specified hierarchy,
identify the CT execution time, inspect the relevant post-CT creatinine
trajectory, check chronic dialysis and ESRD (end-stage renal disease) status, apply the KDIGO \cite{kdigo2012aki} creatinine criteria, and cite the record evidence supporting the resulting
verdict. The evaluation needs to inspect whether the agent retrieved the necessary evidence,
used the appropriate temporal window and clinical rule, and avoided unsupported shortcuts. The workflow illustrates why professional-work evaluation
may need to measure more than final-answer correctness alone.

The workflow evaluator, $g_w$, measures multiple aspects of the observable
execution, including verdict correctness, appropriate abstention, evidence
grounding, policy grounding, and adherence to scenario-specific investigation
requirements. The workflow-success predicate, $\phi_w$, specifies which of
these measurements are required for an execution to count as successful. For
example, an illustrative process-aware success definition may take the form

\begin{equation}
u_i
=
\mathbf{1}
\left\{
\text{verdict correct}
\;\land\;
\text{no disqualifying process defect}
\right\}.
\label{eq:clinicare-deployment-success}
\end{equation}

The clinical-audit case study uses a terminal accept-or-escalate policy class.
Here, the agent's stated confidence $s_i$ is recorded as the policy-observable signal, and
the oversight policy is defined based on how that signal is used to route a
completed adjudication between autonomous acceptance and human review. Confidence is therefore a
property of this particular policy instantiation, not a requirement of the
general \EFE{} framework.

Because the routing decision occurs only after the agent has completed the
audit, changing the policy does not change the underlying agent trajectory
$Z_i$. The execution evidence is trajectory-invariant, allowing multiple
candidate oversight policies to be evaluated from the same saved runs.

Given a reliability target and deployment assumptions about human-review
effectiveness and operating cost, \EFE{} uses this evidence to select an
oversight policy and then statistically qualify the frozen operating point.
The clinical-audit workflow provides a concrete instance of the
trajectory-invariant policy-evaluation setting developed more fully in
Section~\ref{sec:qualification}.

%% file: sections/metrics.tex
\section{Oversight Policy Optimization and Deployment Qualification}
\label{sec:qualification}

Section~\ref{sec:overview} formulated reliable deployment as a constrained optimization problem over oversight policies. We now develop the quantities, policy-optimization procedures, and statistical qualification methods needed
to operationalize that formulation.

The key distinction is between \emph{agent performance} and
\emph{deployed-system reliability}. An oversight policy determines how autonomous agent execution is combined with human intervention. Its value depends not only on how often the agent succeeds on its own, but also
on how effectively the policy uses observable workflow information to introduce oversight, how effective that oversight is, and what it costs.

We first develop the trajectory-invariant terminal accept-or-escalate setting, where many candidate policies can be evaluated from the same saved execution evidence. We then describe statistical qualification of a selected policy on
held-out cases and extend the formulation to trajectory-dependent oversight, where intervention changes the execution trajectory itself.

\subsection{Qualification Quantities}
\label{sec:qualification-quantities}

For agent $m$, workflow $w$, and oversight policy $\pi$, let
$u^\pi \in \{0,1\}$ denote the workflow-specific success event defined in Section~\ref{sec:overview-setting}. The event $u^\pi$ records whether an individual workflow execution under policy $\pi$ satisfies the pre-specified success criterion.

We define the reliability of the human--AI system as

\begin{equation}
R_{m,w}(\pi)
=
\mathbb{E}_{x\sim \mathcal{D}_w}
\left[
u^\pi(x)
\right],
\label{eq:deployment-reliability}
\end{equation}
where \(u^\pi(x)\in\{0,1\}\) denotes whether execution on workflow instance
\(x\) satisfies the workflow-specific success criterion. The expectation is
over workflow instances \(x\) drawn from the population represented by
\(\mathcal{D}_w\) and, when execution is stochastic, over the induced randomness in
agent execution, the environment, oversight process, and interacting actors.
Since \(u^\pi(x)\) is binary, \(R_{m,w}(\pi)\) is the probability that an
execution under policy \(\pi\) satisfies the workflow-specific success criterion on a randomly drawn workflow instance. Deployment qualification is an aggregate statistical claim about the human--AI system induced by $\pi$, not a property assigned to any individual task instance.

Let $K_w(Z^\pi)$ denote the workflow-specific operating cost associated with an execution under policy $\pi$. This may include agent execution as well as human review and intervention. The expected operating cost is

\begin{equation}
C_{m,w}(\pi)
=
\mathbb{E}\!\left[K_w(Z^\pi)\right].
\label{eq:deployment-cost}
\end{equation}

Conceptually, this cost can be decomposed into agent and human components, where both terms may depend on $\pi$, as intervention may increase human effort while changing the subsequent agent execution.

\begin{equation}
C_{m,w}(\pi)
=
C_{\mathrm{agent}}(m,w,\pi)
+
C_{\mathrm{human}}(m,w,\pi),
\label{eq:cost-decomposition}
\end{equation}

Equations~\ref{eq:deployment-reliability} and
\ref{eq:deployment-cost} define the two canonical quantities in the
deployment objective of Equation~\ref{eq:efe-overview-objective}. They are
properties of the deployed human--AI configuration under a specified policy,
not intrinsic properties of the agent alone.

\subsection{Terminal Accept-or-Escalate Oversight}
\label{sec:accept-escalate}
We first consider the trajectory-invariant terminal accept-or-escalate setting used in our clinical-audit case study. In this setting, the agent completes its work before an oversight decision is made. The completed execution provides a result together with an observable routing signal $s_i$; the policy then determines whether the result is accepted autonomously or routed to a declared human-review path.

To evaluate such policies, \EFE{} first executes the agent on representative
workflow instances under autonomous execution. For each task instance $x_i$,
this produces a saved trajectory $Z_i$, an observable routing signal $s_i$,
and an autonomous-success indicator $u_i \in \{0,1\}$.

Here, $(Z_i,s_i,u_i)$ are evaluation data used for policy qualification;
at deployment time, the routing decision is made from observable information
such as $s_i$, while $u_i$ is generally unknown.

Because the terminal routing decision occurs only after $Z_i$ has been generated, the trajectory is invariant to the candidate routing policy. \EFE{} can evaluate many accept-or-escalate policies by replaying the same saved evaluation runs instead of re-executing the agent for every candidate policy.

For a scalar routing signal $s_i$, consider a threshold policy $\pi_\tau$
that accepts the completed result autonomously when $s_i \geq \tau$ and
routes it to human review otherwise.
We assume that larger values of $s_i$ indicate greater likelihood of
successful autonomous execution, so that $s_i$ provides a meaningful ordering
of cases for selective routing. The formal routing assumption is given in
Appendix~\ref{app:routing-monotonicity}. The autonomous coverage under threshold $\tau$ is

\begin{equation}
c(\tau)
=
\Pr(s \geq \tau),
\label{eq:coverage}
\end{equation}

and its corresponding human-review rate is

\begin{equation}
e(\tau)
=
1-c(\tau).
\label{eq:review-rate}
\end{equation}

Coverage is useful in this terminal setting because it is simply the
complement of review burden; it is not required as a universal \EFE{}
deployment metric. Among autonomously accepted cases, define

\begin{equation}
a(\tau)
=
\Pr(u=1 \mid s\geq\tau),
\label{eq:selective-success}
\end{equation}

the success probability of work accepted without review.

Let $a_h(\tau)$ denote the probability that a case routed by policy
$\pi_\tau$ is successfully resolved by the declared human-review path.
The reliability of the deployed human--AI system is then

\begin{equation}
R(\tau)
=
c(\tau)a(\tau)
+
\bigl(1-c(\tau)\bigr)a_h(\tau).
\label{eq:terminal-reliability}
\end{equation}

The first term corresponds to work accepted autonomously; the second
corresponds to work routed to human review.

The review-path performance $a_h(\tau)$ can be either measured from the review process or introduced as a deployment assumption.
It is important to note that because routing deliberately selects a non-random subset of cases, $a_h(\tau)$ may vary with the policy: cases routed under a more selective policy may be systematically more difficult than cases routed under another policy.
When a constant review-success assumption $a_h$ is available,
Equation~\ref{eq:terminal-reliability} reduces to

\begin{equation}
R(\tau)
=
c(\tau)a(\tau)
+
\bigl(1-c(\tau)\bigr)a_h.
\label{eq:terminal-reliability-constant-review}
\end{equation}

This formulation makes clear why autonomous benchmark performance does not
determine deployment reliability. Two agents with similar autonomous
success rates can support substantially different human--AI operating points
because the oversight policies available to them may differ in how effectively
they identify cases requiring intervention.

\subsection{Cost at a Target Reliability}
\label{sec:cost-reliable-deployment}

In terminal accept-or-escalate deployment, the agent has already executed
before the routing decision is made. Let $k_m$ denote its expected execution
cost per case and let $k_h$ denote the incremental cost of the declared
human-review path. Under a constant-cost approximation,

\begin{equation}
C(\tau)
=
k_m
+
k_h\,e(\tau)
=
k_m
+
k_h\bigl(1-c(\tau)\bigr).
\label{eq:terminal-cost}
\end{equation}

For target reliability $Y$, the population-level policy optimization problem is

\begin{equation}
\tau^\star(Y)
=
\arg\min_{\tau}
C(\tau)
\qquad
\text{subject to}
\qquad
R(\tau)\geq Y,
\label{eq:threshold-optimization}
\end{equation}

together with any additional workflow-specific risk constraint.

Equation~\ref{eq:threshold-optimization} defines the ideal population
operating point. In practice, $R(\tau)$ and $C(\tau)$ are unknown and policy selection is performed using development evidence. Let
$\widehat{R}_{\mathrm{dev}}(\tau)$ and
$\widehat{C}_{\mathrm{dev}}(\tau)$ denote the corresponding development-set estimates. A generic development-stage selection rule takes the form

\begin{equation}
\widehat{\tau}
\in
\arg\min_{\tau}
\widehat{C}_{\mathrm{dev}}(\tau)
\qquad
\text{subject to}
\qquad
\widehat{R}_{\mathrm{dev}}(\tau)
\geq
Y_{\mathrm{sel}},
\label{eq:empirical-threshold-selection}
\end{equation}

where $Y_{\mathrm{sel}}$ is a pre-specified development-stage selection
criterion. It may equal the deployment target $Y$ or incorporate a
pre-specified margin or other conservative selection rule.

\EFE{} does not require a particular optimization algorithm or
development-stage selection rule. What is required for qualification is that
all policy choices be made without using the held-out qualification cases.
The selected policy $\widehat{\pi}$ is subsequently frozen and evaluated
against the actual deployment target $Y$ as described in
Section~\ref{sec:statistical-qualification}.

When $k_m$ and $k_h>0$ are constant, minimizing operating cost is equivalent
to minimizing human-review burden, or equivalently maximizing autonomous
coverage, among feasible policies. Writing

\begin{equation}
e^\star(Y)
=
e\!\left(\tau^\star(Y)\right),
\end{equation}

the population cost of reliable deployment is

\begin{equation}
C^\star(Y)
=
k_m+k_h e^\star(Y).
\label{eq:cost-of-reliable-deployment}
\end{equation}

This provides an economically meaningful basis for comparing systems:
how much operating cost is required for each system to satisfy the same
workflow-specific reliability requirement.

The constant-cost model is useful for exposition but is not required by
\EFE{}. Review effort may vary substantially across cases, and different
policies may trigger different forms of review, correction, or escalation.
When case-level costs are available, \EFE{} estimates $C(\pi)$ directly from
the costs induced by the candidate policy instead of assigning a single
constant $k_h$ to every reviewed case.

\subsection{Quality of the Routing Signal}
\label{sec:routing-quality}

For terminal accept-or-escalate policies, the value of oversight depends in part on whether the routing signal identifies cases for which autonomous
execution is likely to fail. A useful signal should rank higher-risk cases
below lower-risk cases, allowing a more selective policy to remove failures
from autonomous handling preferentially.

A natural diagnostic is the selective-risk curve,
\begin{equation}
r(\tau)
=
\Pr(u=0 \mid s\geq\tau),
\label{eq:selective-risk}
\end{equation}
or equivalently $r(c)$ when operating points are indexed by autonomous
coverage. Sweeping the routing threshold traces the risk--coverage
relationship: a stronger routing signal achieves lower selective risk at the
same autonomous coverage.

We summarize this curve using the area under the risk--coverage curve (\textbf{AURC}),
\begin{equation}
\mathrm{AURC}
=
\int_0^1 r(c)\,dc,
\label{eq:aurc}
\end{equation}
with lower values indicating a more favorable risk--coverage
tradeoff. 

AURC is influenced both by routing quality and by the base autonomous success rate $a_0=\Pr(u=1)$. We also report the \textbf{AUROC} of the routing signal $s$ for predicting execution success $u$. AUROC measures ranking discrimination and, unlike AURC, is not directly determined by the prevalence of successful executions.

These metrics are diagnostics for the terminal routing policy class rather
than universal \EFE{} qualification quantities. Their practical importance is
ultimately determined by the operating policies they support: at a fixed
reliability target, better routing can reduce the amount of human intervention, and the operating cost, required for qualification.

\subsection{Statistical Qualification}
\label{sec:statistical-qualification}

Optimizing an oversight policy on development evidence does not establish that
the selected policy will satisfy its deployment requirements on future cases.
\EFE{} separates \emph{policy selection} from
\emph{statistical policy qualification}.

The evaluation sample $S_w$ is partitioned by disjoint index sets
$I_{\mathrm{dev}}, I_{\mathrm{qual}} \subseteq [N_w]$ into a development set
$S_{\mathrm{dev}} = \{x_i : i \in I_{\mathrm{dev}}\}$ and a held-out
qualification set $S_{\mathrm{qual}} = \{x_i : i \in I_{\mathrm{qual}}\}$. All choices that determine the policy---including routing rule, threshold, optimization procedure, and any development-stage
margin---are made using $S_{\mathrm{dev}}$. The selected policy
$\widehat{\pi}$ is then frozen before the qualification outcomes are examined.

For terminal accept-or-escalate oversight, let
$\widehat{\tau}$ denote the threshold selected on the development set.
For each qualification case $i$, define

\begin{equation}
d_i
=
\mathbf{1}\{s_i \geq \widehat{\tau}\},
\label{eq:qualification-routing}
\end{equation}

where $d_i=1$ denotes autonomous acceptance and $d_i=0$ denotes human review. The qualification set is used only to estimate the operating characteristics
of this fixed policy, not to search for a better threshold.

\paragraph{Measured human-review outcomes.}

When the outcome of the declared human-review path is observed for each
escalated case, let $h_i\in\{0,1\}$ denote whether that review path produces
a successful result.
The realized success of the deployed system on qualification case $i$ is

\begin{equation}
v_i
=
d_i u_i + (1-d_i)h_i,
\label{eq:qualification-case-success}
\end{equation}

and the empirical deployment reliability is

\begin{equation}
\widehat{R}(\widehat{\pi})
=
\frac{1}{N_{\mathrm{qual}}}
\sum_{i\in I_{\mathrm{qual}}} v_i,
\qquad
N_{\mathrm{qual}} = |I_{\mathrm{qual}}|.
\label{eq:qualification-reliability}
\end{equation}

In the terminal setting, $v_i$ is the realized policy-level success outcome
corresponding to the general success event $u_i^\pi$ in
Section~\ref{sec:qualification-quantities}; $u_i$ here denotes the success
of the autonomous execution before routing.

When every policy-level outcome is observed and qualification cases are
independent, $v_i$ is binary and a one-sided binomial lower confidence bound can be computed directly. At confidence level $1-\alpha$, \EFE{} qualifies
the frozen policy only if

\begin{equation}
R_{\mathrm{LCB}}^{\,1-\alpha}
(\widehat{\pi})
\geq Y.
\label{eq:reliability-lcb}
\end{equation}

A point estimate above the target is not sufficient: the held-out evidence must support a reliability lower bound that itself clears the deployment requirement.

\paragraph{Human-review performance as a deployment assumption.}

In retrospective benchmark analysis, the declared human-review path may not
be executed for every escalated case. Its effectiveness can instead be
introduced explicitly as a deployment assumption. Suppose, for example, that every routed case is assumed to be successfully
resolved with probability $a_h$. For each qualification case define the
policy-level conditional success value

\begin{equation}
y_i(a_h)
=
d_i u_i
+
(1-d_i)a_h
\in[0,1].
\label{eq:qualification-case-success-assumed-human}
\end{equation}

Under the constant-review assumption, the population reliability of the fixed
policy is

\begin{equation}
R(\widehat{\pi};a_h)
=
\mathbb{E}
\left[
y_i(a_h)
\right],
\end{equation}

This quantity is no longer a simple binomial proportion, because reviewed cases contribute an assumed success probability rather than an observed
binary outcome. Writing $n_{\mathrm{acc}}$ for the number of accepted
qualification cases and $\widehat{p}_{\mathrm{acc}}$ for the observed success
rate among them, the empirical estimate can be written as

\begin{equation}
\widehat{R}(\widehat{\pi};a_h)
=
\frac{n_{\mathrm{acc}}}{N_{\mathrm{qual}}}\,
\widehat{p}_{\mathrm{acc}}
+
\frac{N_{\mathrm{qual}}-n_{\mathrm{acc}}}{N_{\mathrm{qual}}}\,
a_h,
\label{eq:qualification-reliability-decomposition-second}
\end{equation}

in which only $\widehat{p}_{\mathrm{acc}}$ is estimated from observed
outcomes.

For statistical qualification, however, both accepted-case success and the
fraction of cases routed by the frozen policy vary across samples from the
workflow population. \EFE{} constructs a one-sided lower confidence
bound directly for the mean of the bounded case-level quantities
$\{y_i(a_h)\}_{i\in I_{\mathrm{qual}}}$ rather than treating the observed
routing fraction as fixed. Denote this bound by

\begin{equation}
R_{\mathrm{LCB}}^{\,1-\alpha}
(\widehat{\pi};a_h).
\end{equation}

The policy qualifies conditionally on the declared human-review assumption
only if

\begin{equation}
R_{\mathrm{LCB}}^{\,1-\alpha}
(\widehat{\pi};a_h)
\geq Y.
\label{eq:reliability-lcb-assumed-human}
\end{equation}

The lower bound may be computed using a pre-specified valid procedure for the
mean of bounded case-level outcomes. When cases are clustered, the
qualification procedure operates at the corresponding cluster level as
described below.

No statistical confidence is attributed to the assumed term.
The qualification statement is explicitly conditional,
of the form ``qualified at target $Y$ under the declared review assumption
$a_h$.''
For policy families whose contributions do not admit this decomposition,
the same conditional bound is obtained by resampling qualification cases and
re-evaluating the fixed policy on each resample.
If $a_h$ is itself estimated from separate human-review data, uncertainty in
that estimate must also be propagated rather than treating $a_h$ as known
exactly; a conservative qualification replaces $a_h$ with a one-sided lower
confidence bound $a_{h,\mathrm{LCB}}$ in
Equation~\ref{eq:reliability-lcb-assumed-human}.

\paragraph{Break-even review assumption.}
Because qualification reliability $R_{\mathrm{LCB}}^{\,1-\alpha}(\widehat{\pi};a_h)$ is nondecreasing
in $a_h$, the dependence on the human-review assumption can be summarized by a single quantity: the weakest review effectiveness under which the frozen policy still qualifies:

\begin{equation}
a_h^{\min}(\widehat{\pi},Y)
=
\inf
\left\{
a_h\in[0,1]
:
R_{\mathrm{LCB}}^{\,1-\alpha}(\widehat{\pi};a_h)
\geq
Y
\right\}.
\label{eq:break-even-review-assumption}
\end{equation}

The deployment profile can then state an auditable requirement on the
human process: the selected policy qualifies at target $Y$ provided the
declared review path achieves effectiveness at least $a_h^{\min}$.

If no $a_h\in[0,1]$ satisfies the criterion, the policy is reported as not
qualified at target $Y$ under any constant review-success assumption.

\paragraph{Sampling unit.}

The statistical sampling unit must match the unit over which the deployment
claim is intended to generalize. Multiple stochastic executions of the same
task instance are not treated as independent new workflow cases. When repeated
runs are present, they are carried together at the task-instance level.

Similarly, when cases are naturally clustered by patient, customer, account,
site, or another higher-level unit, qualification must preserve that
dependence structure, for example by constructing confidence intervals or
resampling at the cluster level.

A policy that fails the statistical criterion is reported as
\emph{not qualified at target $Y$} or as having \emph{insufficient evidence};
it is not retuned using the qualification set. Any change to the policy $\widehat{\pi}$ after
examining qualification outcomes requires a new held-out qualification
evaluation.

Unless otherwise stated, \EFE{} qualification claims are \emph{pointwise} for a pre-specified deployment target and frozen policy. Simultaneous statistical claims across multiple targets, policies, or operating points require an appropriate simultaneous-inference procedure.

\subsection{Risk-Sensitive Qualification}
\label{sec:risk-sensitive-qualification}

Reliability treats every unsuccessful execution as the same binary event. For
workflows in which failures differ materially in consequence, \EFE{} may
additionally define a workflow-specific deployment loss

\begin{equation}
L_w(Z^\pi)\geq 0
\end{equation}

and impose the constraint

\begin{equation}
\rho_w\!\left(L_w(Z^\pi)\right)
\leq B_w,
\label{eq:risk-qualification}
\end{equation}

where $B_w$ is the declared risk tolerance and $\rho_w$ is a
workflow-specific risk functional. Examples include a bound on the
probability of a specified severe event or a tail-risk measure such as
Conditional Value at Risk (CVaR).

This constraint is optional. The common \EFE{} objective remains reliable
deployment at minimum operating cost; an additional risk constraint is
introduced when binary success alone does not adequately represent the
consequences of failure. Estimation and statistical qualification of
risk-sensitive constraints are discussed further in Appendix~\ref{app:tail-risk-qualification}.

\subsection{Trajectory-Dependent Oversight}
\label{sec:trajectory-dependent-oversight}

The terminal formulation above is especially convenient because the routing
decision does not change the saved agent trajectory.
More general enterprise workflows allow intervention during execution.

Let $h_t$ denote the observable workflow history at decision point $t$, and
let

\begin{equation}
\pi(o_t\mid h_t)
\end{equation}

denote an oversight policy over intervention decisions $o_t$, such as
approval, correction or takeover.
Executing the workflow under $\pi$ induces a policy-dependent trajectory

\begin{equation}
Z_i^\pi
\sim
P(\,\cdot\mid m,x_i,E_w,\pi\,).
\label{eq:policy-dependent-trajectory}
\end{equation}

The qualification objective remains the one defined in
Equation~\ref{eq:efe-overview-objective}.
The difference is how the evidence needed to estimate that objective is generated.

For trajectory-invariant terminal policies, candidate policies can be
evaluated by replaying saved trajectories under different routing rules. For a
trajectory-dependent policy, changing $\pi$ changes what the agent, oversight
actor, or interacting user subsequently observes and does. Reliability, cost,
and risk must instead be estimated by executing or appropriately simulating
the candidate policy in the evaluation runtime.

The phases introduced in Section~\ref{sec:overview-flow} should therefore be
understood as a logical evaluation flow rather than necessarily a one-pass
computation. For trajectory-dependent oversight, policy optimization may
repeatedly invoke agent execution and workflow evaluation while searching over
candidate policies. 

Policy-in-the-loop evaluation may change both sides of the deployment
tradeoff. Intervention can improve the probability of successful completion,
but can also change agent execution length, human effort, recovery behavior,
and the distribution of subsequent outcomes. These effects are captured by
the same general quantities $R_{m,w}(\pi)$, $C_{m,w}(\pi)$, and, when
applicable, $\rho_w(L_w(Z^\pi))$.

This setting is naturally connected to simulation-based and black-box
optimization, where objective and constraint values need not admit closed-form
expressions and can instead be estimated through stochastic simulation or
system execution~\cite{amaran2016simulation,audet2017derivativefree}. \EFE{} does not prescribe an optimization algorithm. The policy representation and optimization method may be
workflow-dependent. What is shared across workflows is the deployment
objective and qualification procedure: candidate policies are evaluated in
terms of the reliability, cost, and risk of the resulting human--AI system,
a policy is selected using development evidence, and the frozen policy is
qualified on held-out evidence.

\paragraph{Empirical instantiation.}
Section~\ref{sec:empirical} applies the qualification methodology above to
CliniCARE-Bench.
The analysis is conducted on the benchmark's full four-way adjudication space
using a signal-based accept-or-escalate policy that is agnostic to the class
labels, with sensitivity analyses in
Appendix~\ref{app:clinicare-supplementary} and a cost-of-conflation control in
Appendix~\ref{app:conflation-control}.

%% file: sections/standard.tex
\section{\EFE{} Evaluation Runtime and Open Testbed}
\label{sec:platform}

Sections~\ref{sec:overview} and~\ref{sec:qualification} define what \EFE{}
qualifies and how an oversight policy is selected.
This section describes the execution and interface testbed that produces the
evidence required by that methodology.

\EFE{} does not introduce a new general-purpose agent-evaluation harness.
Its reference implementation uses Inspect AI~\cite{inspect_ai} for task
execution, agent integration, sandboxed environments, scoring, intervention,
and evaluation logging. \EFE{} adds the workflow-level semantics and deployment-qualification layer needed to translate executions into a reliability--oversight--cost deployment profile.

\subsection{Runtime Layering and Reference Execution Substrate}
\label{sec:platform-layering}

The implementation follows the three-layer architecture in
Figure~\ref{fig:efe_overview}.
Non-executing \emph{specification artifacts} define the workflow,
evaluation semantics, and deployment assumptions.
The \emph{evaluation runtime} instantiates these specifications and executes
the agent, workflow environment, and any required oversight or user actors.
The resulting execution evidence is evaluated and passed to the qualification
methodology in Section~\ref{sec:qualification}.

\EFE{} standardizes the semantics and evidence that cross these boundaries
without requiring every evaluation to share a task representation, agent
implementation, environment technology, or grading mechanism, so existing
evaluation infrastructure can be reused. 

In the Inspect reference implementation, an executable evaluation is
represented by an Inspect \texttt{Task}, which combines a collection of
\texttt{Sample}s with an agent or \texttt{Solver}, one or more \texttt{Scorer}s, and
the required runtime configuration.
\EFE{} introduces the workflow and deployment semantics above these execution
objects. The correspondence is an implementation mapping, not a one-to-one semantic equivalence.

\subsection{Specification Interfaces}
\label{sec:platform-specifications}

The runtime consumes three logically distinct classes of specification
artifacts corresponding to Figure~\ref{fig:efe_overview}.

\paragraph{Workflow specification.}
The workflow specification identifies the reusable workflow, $w$, the
task-instance population, $\mathcal{D}_w$, from which the evaluation sample,
$S_w$, is drawn, and the environment specification, $E_w$.
It defines the work to be performed and the population for which the
evaluation provides evidence.
A concrete task instance, $x_i$, supplies the case-specific inputs for one
execution.

\EFE{} does not require a second executable task format.
In the Inspect reference implementation, a \EFE{} task instance is represented
by an Inspect \texttt{Sample}, and a collection of task instances is
represented by an Inspect \texttt{Dataset}.
The enclosing Inspect \texttt{Task} specifies how those samples are executed
and evaluated.
This distinction is important: an \EFE{} \emph{task instance} corresponds to an
Inspect \texttt{Sample}, not to an Inspect \texttt{Task}.

\paragraph{Evaluation specification.}
The evaluation specification defines the workflow evaluator, $g_w$, the
workflow-success predicate, $\phi_w$, and any routing signal exposed to an
oversight policy.

It specifies the transformation
$Z_i^\pi \xrightarrow{g_w} \mathbf{q}_i^\pi \xrightarrow{\phi_w} u_i^\pi$
of Equation~\ref{eq:evaluation-to-deployment}.

The mechanisms used to compute $\mathbf{q}_i^\pi$ may be deterministic,
human-annotated, model-assisted, or composed from multiple evaluation
mechanisms.
In Inspect, these computations can be implemented through one or more
\texttt{Scorer}s, including externally computed scores when appropriate.
\EFE{}'s evaluator, $g_w$, denotes the workflow-level semantics of what is
measured; an Inspect \texttt{Scorer} is one executable mechanism for
producing those measurements.
The success predicate, $\phi_w$, remains an \EFE{}-level definition that determines
which measurements constitute acceptable execution for deployment
qualification.

\paragraph{Deployment assumptions.}
Deployment assumptions remain separate from measured evaluation evidence.
They include the reliability target, $Y$, the human-oversight model,
operating-cost assumptions, and any workflow-specific risk tolerance.
Different organizations may therefore interpret the same measured execution
evidence under different deployment requirements.

The development and held-out qualification partitions required by
Section~\ref{sec:statistical-qualification} are versioned with the \EFE{}
evaluation configuration.
Instances used to select an oversight policy must remain separate from the
instances used to qualify the frozen policy.

Table~\ref{tab:efe-runtime-interface} summarizes the minimal logical
interface required of an \EFE{}-compatible runtime and its reference
realization in Inspect.

\begin{table}[ht!]
\centering
\small
\setlength{\tabcolsep}{8pt}
\renewcommand{\arraystretch}{1.4}
\renewcommand{\tabularxcolumn}[1]{m{#1}}
\rowcolors{2}{white}{scalePanel!35}
\begin{tabularx}{\textwidth}{@{}
  >{\raggedright\arraybackslash}m{2.9cm}
  Y
  >{\raggedright\arraybackslash}m{4.0cm} @{}}
\toprule
\rowcolor{scalePanel!75}
\textbf{\EFE{} interface}
& \textbf{Required semantics}
& \textbf{Inspect reference realization}
\\
\midrule

\textbf{Task population}
& Represent the sampled workflow instances $S_w$, $x_i \sim \mathcal{D}_w$,
with stable identifiers and required case inputs.
& \texttt{Dataset} of \texttt{Sample}s
\\

\textbf{Agent execution}
& Execute agent $m$ on $x_i$ under the declared model, harness, and runtime
configuration.
& \texttt{Agent} or \texttt{Solver}; external agents through Agent Bridge
\\

\textbf{Workflow environment}
& Expose the tools, data, files, systems, and computational state specified by
$E_w$.
& Task/sample tools and sandbox configuration
\\

\textbf{Oversight interaction}
& Allow policy $\pi$ to approve, reject, clarify, intervene, escalate, or invoke
another actor when the workflow requires it.
& Approval policies, approver, agent intervention, or task-specific actors
\\

\textbf{Execution evidence}
& Preserve the observable evidence required to interpret $Z_i^\pi$, including
the final work product and relevant interaction history.
& Per-sample evaluation log, transcript, outputs, metadata, and retained
artifacts
\\

\textbf{Workflow evaluation}
& Compute $\mathbf{q}_i^\pi=g_w(Z_i^\pi)$ and retain the measurements used by
the deployment criterion.
& One or more \texttt{Scorer}s or externally supplied scores
\\

\textbf{Qualification inputs}
& Expose $u_i^\pi$, routing signal $s_i$, resource/cost measurements, and the
sampling identifiers required by \EFE{} qualification.
& Scores, sample metadata, transcript/log fields, usage measurements, and
\EFE{}-derived records
\\

\textbf{Evaluation lineage}
& Identify the workflow, case, agent, model, environment, evaluator, policy,
and execution configuration associated with the evidence.
& \texttt{Task}/\texttt{Sample} configuration and versioned evaluation logs
\\

\bottomrule
\end{tabularx}

\caption{
\textbf{Minimal logical interface for an \EFE{}-compatible evaluation runtime.}
\EFE{} defines the semantics required for deployment qualification; the rightmost
column shows their reference realization in Inspect AI.
Other execution substrates may implement the same interface using different
native abstractions.
}
\label{tab:efe-runtime-interface}
\end{table}

\subsection{Execution, Actors, and Policy Evaluation}
\label{sec:platform-execution}

The evaluation runtime instantiates the workflow environment and executes
agent $m$ on task instance $x_i$ under oversight policy $\pi$.
The runtime is responsible for making the workflow-specified tools, data,
systems, and interaction interfaces available while preserving the declared
evaluation conditions.

Inspect provides the reference execution mechanisms for this layer.
Agents may be implemented through native Inspect agents or solvers, or
integrated from external agent frameworks through the Inspect Agent Bridge.
Workflow-specific files and computational environments can be attached to
individual samples and executed in controlled sandboxes.
These implementation choices do not change the \EFE{} semantics of the workflow
or trajectory.

For workflows requiring interaction, \EFE{} distinguishes two additional actor
roles.
An \emph{oversight actor} represents an organization-side reviewer or
operator who may approve, clarify, correct, reject, escalate, or take over
work.
A \emph{user actor} represents an external user or counterparty interacting
with the deployed system.
Either actor may be human or simulated, but its identity, configuration, and
available actions must be part of the evaluation configuration.

Inspect approval policies and agent-intervention mechanisms provide reference
implementations for several forms of oversight.
For example, approval policies can require selected tool calls to be approved,
rejected, or escalated, while agent intervention permits a human operator to
interrupt or redirect an executing agent.
These mechanisms do not define the \EFE{} policy class $\Pi_w$; they are runtime
primitives through which particular policies may be implemented.
Other forms of interaction may be realized through custom agents, solvers, or
workflow-specific simulators.

The runtime supports both policy-evaluation procedures introduced in
Section~\ref{sec:overview-flow}.

For \emph{trajectory-invariant} policies, the relevant agent execution is
independent of the terminal routing decision.
The runtime can therefore execute the agent once, preserve the resulting
evidence, and evaluate multiple candidate policies by replaying the same
completed executions.
The accept-or-escalate analysis in Section~\ref{sec:accept-escalate} uses
this path.

For \emph{trajectory-dependent} policies, intervention changes what happens
next.
The candidate policy must participate in execution, and the runtime records
the resulting policy-dependent trajectory $Z_i^\pi$.
Reliability, cost, and risk for that policy must be estimated from
executions or simulations performed under the policy itself.

\subsection{Evaluation Evidence and Workflow Evaluation}
\label{sec:platform-evidence}

Each execution produces a sample-level evaluation record containing the
observable evidence required by the workflow evaluator and qualification
analysis.
\EFE{} does not introduce a separate universal trajectory format.
Instead, $Z_i^\pi$ denotes the workflow-relevant observable execution
trajectory represented by the underlying runtime record.

In the Inspect reference implementation, evaluation logs preserve
sample-level outputs, model and tool interactions, scoring information,
metadata, usage information, and the execution transcript.
For bridged agents, model interactions are likewise captured in the Inspect
transcript.
When human or automated interventions occur during execution, those
interventions should remain part of the retained evidence.

The workflow evaluator applies $g_w$ to this record to produce
$\mathbf{q}_i^\pi$, and the pre-specified predicate $\phi_w$ derives the
corresponding success event $u_i^\pi$.
Diagnostic measurements remain available independently of this binary
success event.
For example, a workflow may retain outcome correctness, evidence grounding,
process adherence, appropriate abstention, and resource use even when only a
subset of those measurements enters $\phi_w$.

Inspect's scoring interface supports multiple workflow measurements without
requiring them to be collapsed into a single benchmark score.
Where the saved execution record contains sufficient evidence, an updated
evaluator can also be applied to an existing evaluation log without
re-executing the agent.

For workflows requiring repeated stochastic execution, the same logical task
instance may be executed for multiple epochs.
\EFE{} retains the underlying task-instance identity and epoch-level evidence so
that repeated runs are not treated as independent task instances during
statistical qualification.

\paragraph{Running example: CliniCARE-Bench.}
For CliniCARE-Bench, each case is represented as an evaluation sample whose
execution record contains the completed clinical adjudication, supporting
patient- and policy-evidence citations, the case-level routing signal, and the
observable investigation trajectory, including retrieved evidence, tool
calls, computations, and intermediate artifacts.
The CliniCARE evaluator derives verdict and process measurements from this
record and applies the workflow-specific success predicate before the
resulting evidence is passed to \EFE{} qualification.

\subsection{Qualification Handoff and Deployment Profile}
\label{sec:platform-qualification-handoff}

The boundary between workflow evaluation and deployment qualification is an
evidence interface rather than a new task evaluator.
The qualification layer consumes measured execution evidence together with
declared deployment assumptions and applies the methodology in
Section~\ref{sec:qualification}.

For the trajectory-invariant accept-or-escalate case, the minimal
qualification record for task instance $i$ contains the workflow-defined
success event $u_i$, the observable routing signal $s_i$, and the execution or
resource measurements required by the cost model.
It also retains the identifiers needed to associate the record with its task
instance, statistical sampling unit, evaluated system, and versioned
evaluation configuration.
These quantities can be extracted from Inspect scores, metadata, usage
measurements, and evaluation logs into the \EFE{} qualification dataset.

\paragraph{Versioned qualification record.}
The qualification record has a stable, versioned schema so that a deployment claim can
be reproduced and audited. Each record carries a \texttt{schema\_version}; a
\texttt{lineage} block that version-stamps the workflow, task population, evaluated
system, environment, evaluator $g_w$, success predicate $\phi_w$, and routing-signal
definition; the per-instance \texttt{measurements}
($u_i$, $s_i$, retained diagnostics $\mathbf{q}_i$, and cost); and the identifiers of the
task instance and statistical sampling unit. Deployment assumptions are recorded
separately with the profile, not in the per-instance record, so the same records support
multiple deployment analyses. 

The development and held-out partitions carry an identical schema; only the
\texttt{partition} field differs. Deployment assumptions ($Y$, $a_h$, review cost) are
attached to the resulting profile, keeping measured evidence separate from the
assumptions under which it is interpreted.

Section~\ref{sec:statistical-qualification} then selects the oversight policy
on development data and qualifies the frozen operating point on held-out
evidence.
The qualification computation is intentionally separate from Inspect's task
scoring: Inspect records what happened and how the workflow evaluator scored
it, while \EFE{} determines which deployment policy is supported by that
evidence.

For trajectory-dependent policies, the same logical handoff applies, but the
measurements correspond to trajectories generated under the candidate policy
$\pi$, and the retained evidence includes the policy and actor configuration
needed to interpret $Z_i^\pi$, $u_i^\pi$, and its associated operating cost.

The output is the \emph{deployment profile} defined in
Section~\ref{sec:overview-flow}: a statistically supported operating point
conditional on the workflow, evaluated system, oversight policy, and declared
deployment assumptions, rather than a context-free model score.

\subsection{Contributing and Maintaining Evaluations}
\label{sec:platform-contribution}

New enterprise workflows can be added to \EFE{} without changing the
qualification methodology or the reference execution infrastructure.
A contribution supplies an executable workflow evaluation together with the
specification artifacts of Section~\ref{sec:platform-specifications}: the
workflow identity and represented population, the evaluation semantics $g_w$
and $\phi_w$, any routing signal, the development and held-out partitions, and
the measurements needed for cost analysis.
The same interface serves public benchmarks and enterprise-internal
evaluations; \EFE{} does not require proprietary data or protected references
to be made public, only that the workflow, evaluation, and reported deployment
claim remain explicit.

\paragraph{Validation.}
Interface conformance establishes interoperability; validation checks whether
the resulting evidence supports a meaningful deployment claim.
A reviewer independent of the contributor checks four aspects against
contributor-supplied evidence: \emph{workflow validity} (a recognizable form of
enterprise work sampled from the stated population), \emph{evaluation validity}
($g_w$ and $\phi_w$ reproduce success labels within a pre-registered tolerance
from observable evidence), \emph{execution validity} (reproducible dry runs with
no hidden-reference access or unintended shortcuts), and \emph{qualification
validity} (a frozen development/held-out partition with policy selection never
touching held-out data).
Verdicts are recorded with the evaluation lineage so that acceptance is
reproducible rather than a matter of reviewer opinion.

\paragraph{Versioned evaluation lineage.}
A \EFE{} result is tied to a versioned \emph{evaluation lineage} rather than just a
benchmark or model name.
The lineage pins the workflow and task population, agent and model
configuration, execution environment, evaluator $g_w$ and predicate $\phi_w$,
routing signal, and the evidence used for qualification, through concrete build
identifiers such as commit hashes, dependency and image digests, and the seeds
governing sampling and any split.
A qualification claim then has the form
\[
\text{evaluation lineage} \;+\; \text{deployment assumptions}
\;\longrightarrow\; \text{deployment profile},
\]
separating the evidence that was measured from the assumptions under which it is
interpreted.
Corrections and later releases create new evaluation records rather than
silently reinterpreting earlier profiles.

\paragraph{Maintenance and requalification.}
Because lineage components are versioned separately, different changes have
different consequences:
\begin{itemize}
  \item \textbf{Assumptions change} $\rightarrow$ \emph{recompute}. When only
  deployment assumptions change (target reliability, review model, or cost), the
  agent evidence is unchanged and \EFE{} recomputes the oversight policy and
  profile from existing records.
  \item \textbf{Evaluation changes} $\rightarrow$ \emph{re-evaluate} when
  evidence is sufficient. When $g_w$ or $\phi_w$ changes, preserved trajectories
  may be re-scored, but only when each trajectory records every input the
  revised evaluator reads; otherwise the case is re-executed.
  \item \textbf{Trajectory or population changes} $\rightarrow$ \emph{re-execute}.
  Material changes to the agent, harness, environment, tools, or represented
  population can alter the trajectory or the claim's scope, so the agent is
  rerun and passed through the same qualification procedure as a new evaluation.
\end{itemize}
\EFE{} thus treats deployment qualification not as a permanent property of an
agent but as an evidence-backed claim attached to a specific lineage and
deployment context.

%% file: sections/results.tex
\section{Empirical Evaluation: Deployment Qualification on CliniCARE-Bench}
\label{sec:empirical}

We use CliniCARE-Bench to illustrate the deployment-qualification methodology of
Section~\ref{sec:qualification}. The goal is not to reproduce the full CliniCARE
benchmark, but to measure the quantities that matter for \EFE{}: whether a routing
signal supports selective autonomy, how much human oversight a given reliability
target demands, and whether systems with similar autonomous performance differ in
deployment value. We organize the section around three questions.

\begin{enumerate}
    \item Does the routing signal separate more reliable from less reliable
    autonomous executions? (Section~\ref{sec:empirical-routing})
    \item How does the required human-review burden grow as the reliability target
    increases? (Section~\ref{sec:empirical-reliability-oversight})
    \item Can systems with similar autonomous performance support materially
    different qualified operating points?
    (Sections~\ref{sec:empirical-qualification-results}--\ref{sec:empirical-hero-comparison})
\end{enumerate}

\subsection{Experimental Setup}
\label{sec:empirical-setup}

\paragraph{Adapting CliniCARE-Bench to \EFE{}.}
Adapting this existing dataset required no change to its clinical content, only
expressing it through the interface of Section~\ref{sec:platform}: each adjudication
case becomes a task instance $x_i$ (an Inspect \texttt{Sample}); the patient-scoped
EHR retrieval tools and sandbox become the environment $E_w$; the four-way verdict
grader becomes the evaluator $g_w$ with success predicate $\phi_w$; the report's
stated confidence becomes the routing signal $s_i$; and the fixed case split becomes
the development and qualification partitions. Deployment assumptions ($Y$, $a_h$, cost)
stay separate, so the same evidence supports the sensitivity analyses of
Appendix~\ref{app:clinicare-supplementary}. The four reference verdicts are

\begin{equation}
\mathcal{Y}
=
\{
\mathrm{Yes},
\mathrm{No},
\mathrm{Indeterminate{:}\ Lack\ of\ Data},
\mathrm{Indeterminate{:}\ Medically\ Ambiguous}
\}.
\label{eq:clinicare-fourway-labels}
\end{equation}

As established in Section~\ref{sec:overview-clinicare}, the two indeterminate outcomes
are task-level clinical judgments---a correct indeterminate verdict is itself a
successful execution---and are distinct from the \EFE{} decision to escalate a
completed case for review. Keeping them separate lets abstention be handled inside the
common \EFE{} optimization without treating every indeterminate output as a failure or
a forced escalation.

\paragraph{Success indicators.}
Let $y_i^\star\in\mathcal{Y}$ denote the reference adjudication, $\widehat y_i$ the
agent adjudication, and $s_i$ the routing signal. The primary four-way success
indicator is

\begin{equation}
u_i
=
\mathbf{1}
\{\widehat y_i = y_i^\star\}.
\label{eq:clinicare-fourway-success}
\end{equation}

and a process-aware variant, used as a robustness check in
Appendix~\ref{app:clinicare-supplementary}, additionally requires the absence of a
disqualifying process defect,

\begin{equation}
u_{i,\mathrm{df}}
=
\mathbf{1}
\left\{
\widehat y_i=y_i^\star
\;\land\;
\text{no disqualifying process defect}
\right\},
\label{eq:clinicare-fourway-defectfree-success}
\end{equation}

\paragraph{Systems, cases, and labels.}
We evaluate $16$ systems on all $750$ cases, giving $12{,}000$ system--case runs.
Reference verdicts follow the natural cohort distribution rather than a balanced one:
\textit{Yes} $352$ ($46.9\%$), \textit{No} $248$ ($33.1\%$),
\textit{Indeterminate: Lack of Data} $123$ ($16.4\%$), and
\textit{Indeterminate: Medically Ambiguous} $27$ ($3.6\%$).

\paragraph{Routing signal.}
The signal $s_i$ is the confidence each system states in its own report (required as a
percentage), which we rescale to $[0,1]$ and observe across the full range. It is
present for $11{,}992$ of $12{,}000$ runs; the eight exceptions, seven stating no
confidence and one producing no report, are assigned $s_i=0$ and escalated first.
Systems differ sharply in signal resolution, from $5$ distinct values (Gemini-3.1-Pro)
to $36$ (GPT-5.4-mini). This granularity is itself deployment-relevant: a system with
few distinct values has correspondingly few reachable operating points.

\paragraph{Development and qualification split.}
Cases are partitioned once, uniformly at random with a fixed seed, into
$|S_{\mathrm{dev}}|=375$ and $|S_{\mathrm{qual}}|=375$. The split is over
\emph{cases}, not runs, so every system is developed and qualified on an identical
cohort and all cross-system comparisons are paired. The draw preserves verdict
proportions without stratification ($S_{\mathrm{dev}}$: $172/126/64/13$;
$S_{\mathrm{qual}}$: $180/122/59/14$).

\paragraph{Human-review assumption.}
Because CliniCARE does not execute the declared review path for every routed case, we
treat human-review success $a_h$ as a deployment assumption, with primary value

\begin{equation}
a_h = 0.9,
\label{eq:empirical-human-review-assumption}
\end{equation}

and sensitivity to alternatives in Appendix~\ref{app:clinicare-supplementary}. All
qualification statements are therefore conditional on this assumption, and we also
report the break-even value $a_h^{\min}$
(Equation~\ref{eq:break-even-review-assumption}), the weakest review success under
which a policy still qualifies. The assumption also imposes a ceiling: because
deployment reliability is a convex combination of autonomous and reviewed outcomes and
selective success $1-r(\tau)$ stays below $a_h$ at every coverage level we observe, no
policy can exceed reliability $a_h=0.90$. Targets above $0.90$ are unreachable here not
because of agent capability but because of the assumed quality of the review
path, a structural point a capability-only benchmark cannot express.

\subsection{Escalation Policy}
\label{sec:empirical-policies}

We evaluate a single terminal accept-or-escalate policy that acts only on the routing
signal, selected on $S_{\mathrm{dev}}$ and frozen before qualification. Because the
workflow is treated as a four-way classification with success indicator $u_i$, the
policy is deliberately agnostic to what the classes \emph{mean}: class semantics enter
only through the success predicate $\phi_w$, never through the escalation rule. The
same threshold applies to all four predicted adjudications,
\begin{equation}
\pi_\tau(s_i)
=
\begin{cases}
\mathrm{accept}, & s_i\geq\tau,\\
\mathrm{review}, & s_i<\tau.
\end{cases}
\label{eq:clinicare-fourway-global-policy}
\end{equation}

Fitting $\tau$ to hit $Y$ directly on a sampled development set is prone to selection
bias and tends to fail the stricter held-out lower-bound test. We instead optimize
against a margin-adjusted target,

\begin{equation}
\tau^\star(Y)
=
\arg\min_{\tau}
C(\tau)
\qquad
\text{subject to}
\qquad
\widehat R_{\mathrm{dev}}(\tau;a_h)\geq Y + \delta,
\label{eq:threshold-optimization-with-margin}
\end{equation}

where $\widehat R_{\mathrm{dev}}(\tau;a_h)$ is the development-partition reliability
estimate under the declared review assumption. All CliniCARE policies use a fixed
margin $\delta = 0.05$.

\subsection{Autonomous Performance and Routing Quality}
\label{sec:empirical-routing}

\begin{figure}[ht]
    \centering
    \includegraphics[width=\linewidth]{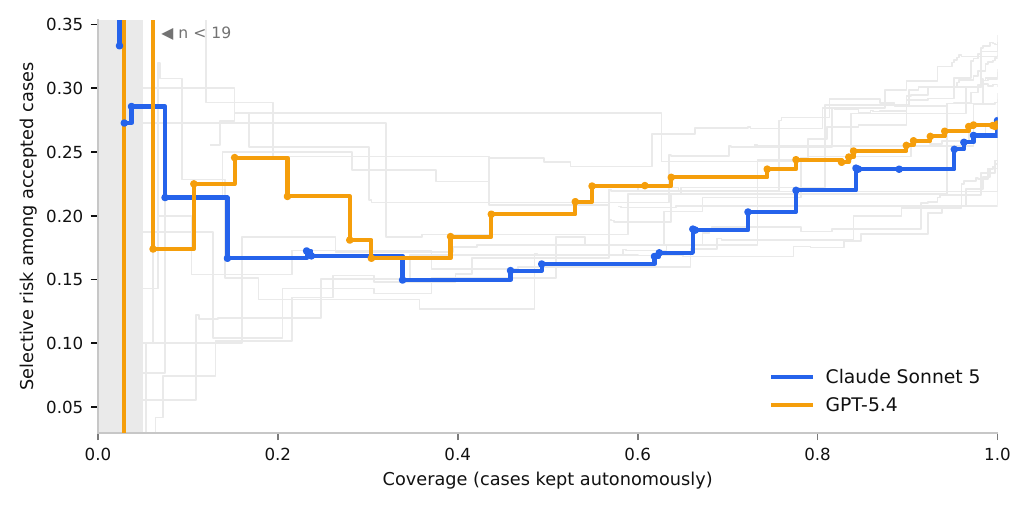}
    \caption{
    \textbf{Risk--coverage behavior of the stated-confidence routing signal
    on the development partition.}
    Sweeping the routing threshold traces the tradeoff between autonomous
    coverage and failure risk among autonomously accepted cases; lower curves
    indicate more effective selective routing.
    Faint curves show all 16 systems; highlighted are the running-example
    pair, Claude Sonnet 5 and GPT-5.4, whose autonomous accuracies differ by
    only 0.3 percentage points. Markers on the highlighted curves denote their reachable
    operating points: only a threshold at a tie-group boundary of a system's
    stated confidences can be selected.
    The shaded band marks coverage levels supported by fewer than 19 accepted
    cases, where the selective-risk estimate is unstable.
    }
    \label{fig:clinicare-riskcov}
\end{figure}

We first ask whether $s_i$ carries information about autonomous success beyond the
aggregate task score. For each system we sweep the threshold
(Equation~\ref{eq:clinicare-fourway-global-policy}) over the development data and
compute the autonomous coverage $c(\tau)$ and selective risk $r(\tau)$ of
Section~\ref{sec:routing-quality}. Figure~\ref{fig:clinicare-riskcov} shows the
resulting risk--coverage curves. An informative signal should lower selective risk as
coverage falls, because the first cases removed from autonomous handling should be
disproportionately failures, which is broadly what we observe. Each curve begins at the
system's lowest reachable coverage; the Gemini models return $100\%$ confidence on up
to $73\%$ of cases, so even their most selective reachable policy still leaves most
runs autonomous.

Table~\ref{tab:clinicare-routing-summary} summarizes autonomous success and routing
quality. Across the $16$ systems, base accuracy $a_0$ and AUROC are essentially
uncorrelated (Pearson $\rho=-0.12$): a system can be accurate yet rank its own outputs
poorly, or the reverse. Qwen-3.7-Plus is the clearest case, pairing one of the lowest
base accuracies with the best AUROC. AURC, by contrast, tracks base accuracy closely
($\rho=-0.80$ with $a_0$), which is exactly why we report both---AURC captures the
absolute risk--coverage tradeoff a deployment faces, while AUROC isolates ranking
quality from the base rate.

\begin{table}[h!]
\centering
\small
\begin{tabular}{lcccc}
\toprule
\textbf{System} & \textbf{$a_0$} & \textbf{AURC $\downarrow$} & \textbf{AUROC $\uparrow$} & \textbf{Succ@50\%} \\
\midrule
\multicolumn{5}{@{}l}{\texttt{Claude Code}}\\
\quad Opus 5 & 75.7 & 0.169 & 0.680 & 83.1 \\
\quad Sonnet 5 & 72.5 & 0.203 & 0.681 & 83.7 \\
\addlinespace
\multicolumn{5}{@{}l}{\texttt{Codex}}\\
\quad GPT-5.6-Sol & 73.1 & 0.189 & 0.640 & 81.8 \\
\quad GPT-5.6-Luna & 69.9 & 0.195 & 0.645 & 79.9 \\
\quad GPT-5.5 & 75.7 & 0.172 & 0.652 & 83.7 \\
\quad GPT-5.4 & 72.8 & 0.222 & 0.601 & 79.2 \\
\quad GPT-5.4-mini & 67.5 & 0.248 & 0.628 & 75.9 \\
\addlinespace
\multicolumn{5}{@{}l}{\texttt{Gemini CLI}}\\
\quad Gemini-3.6-Flash & 71.7 & 0.223 & 0.619 & 79.3 \\
\quad Gemini-3.5-Flash & 70.4 & 0.235 & 0.618 & 78.0 \\
\quad Gemini-3.1-Pro & 70.9 & 0.261 & 0.563 & 74.5 \\
\addlinespace
\multicolumn{5}{@{}l}{\texttt{opencode}}\\
\quad DeepSeek-V4-Pro & 68.5 & 0.229 & 0.664 & 79.2 \\
\quad DeepSeek-V4-Flash & 68.8 & 0.263 & 0.602 & 75.4 \\
\quad GLM-5.2 & 75.7 & 0.168 & 0.652 & 82.8 \\
\quad Qwen-3.7-Plus & 66.4 & 0.219 & 0.711 & 79.8 \\
\quad MiniMax-M3 & 66.4 & 0.255 & 0.676 & 79.8 \\
\quad Kimi-K2.7-Code & 65.9 & 0.261 & 0.661 & 76.1 \\
\bottomrule
\end{tabular}
\caption{\textbf{Autonomous performance and routing quality on the development partition.} $n=375$ cases per system. $a_0$ is four-way verdict accuracy under full autonomy, with no review. AURC summarises the risk--coverage curve of Figure~\ref{fig:clinicare-riskcov} and is the quantity the deployment frontier is built from; lower is better. AUROC measures ranking quality independent of baseline accuracy; higher is better. Succ@50\% is selective success over the half of cases each system is most confident in.}
\label{tab:clinicare-routing-summary}
\end{table}

\subsection{Reliability--Oversight Frontier}
\label{sec:empirical-reliability-oversight}

\begin{figure}[ht]
    \centering
    \includegraphics[width=\linewidth]{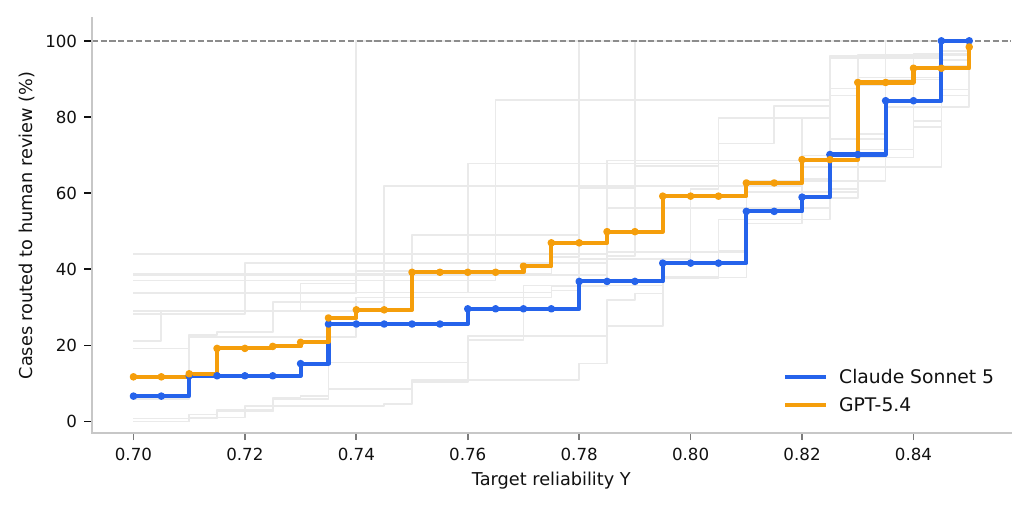}
\caption{
    \textbf{Reliability--oversight frontier on the held-out qualification
    partition}, at review assumption $a_h=0.90$.
    The same routing evidence is converted into the minimum human-review
    burden required to satisfy increasing reliability targets $Y$.
    Policies are selected on the development partition at $Y+\delta$ and
    frozen before qualification.
    Faint curves show all 16 systems; highlighted are the running-example
    pair: at $Y=0.76$, Claude Sonnet 5 qualifies while routing 29.6\% of
    cases to human review against GPT-5.4's 39.2\%, despite near-identical
    autonomous accuracy (72.5\% vs.\ 72.8\%).
    Because $a_h=0.90$ and selective success is below it at every observed
    coverage level, no policy can exceed reliability $0.90$; every system's
    review burden exceeds $90\%$ by $Y=0.85$.}
    \label{fig:clinicare-frontier}
\end{figure}

Routing quality becomes operational when translated into the oversight required to
meet a reliability target. For each target $Y$ we solve
Equation~\ref{eq:threshold-optimization-with-margin} on $S_{\mathrm{dev}}$ to select
the highest-coverage threshold meeting the requirement, freeze it, and apply it to the
held-out $S_{\mathrm{qual}}$. Figure~\ref{fig:clinicare-frontier} plots the resulting
human-review burden against target reliability. This \EFE{} deployment frontier shows
how much work must be routed to review as the organization demands more reliable
operation, and it lets systems be compared at a common deployment requirement rather
than a common coverage level: at a fixed target, the system needing less review
supports the more efficient policy under the same review assumption.

\subsection{Held-Out Qualification and Deployment Comparison}
\label{sec:empirical-qualification-results}

\EFE{} separates a development-set operating point from a statistically qualified
deployment claim: we evaluate each frozen policy on $S_{\mathrm{qual}}$ and compute
the reliability estimate $\widehat{R}$ together with its one-sided $95\%$ lower
confidence bound (LCB) from Section~\ref{sec:statistical-qualification}. Following Equation~\ref{eq:reliability-lcb-assumed-human}, the
LCB attaches exact Clopper--Pearson confidence to the observed accepted component and
\emph{no} confidence to the assumed review term $a_h$; the policy qualifies iff
$\text{LCB}\ge Y$. The step-by-step computation is given in
Appendix~\ref{app:clinicare-lcb-procedure}.
Table~\ref{tab:clinicare-qualification} reports qualification at $Y=0.76$.

The table separates two ways a point-estimate optimum fails to become a deployment.
GLM-5.2 requires the lowest review burden in the table
($10.9\%$) and its point estimate clears the target ($\widehat{R} = 0.762$), yet it
does not qualify. Because it accepts $89\%$ of cases, almost all of its reliability
rests on the measured component, giving it the widest confidence interval in the
table ($\widehat{R} - \mathrm{LCB} = 0.037$), while its point estimate clears $Y$ by
only $0.002$. The same low review burden also leaves the review assumption without
leverage: raising $a_h$ moves only the $10.9\%$ of mass that is reviewed, so no
$a_h \in [0,1]$ rescues it. Making GLM-5.2 qualify would require $\delta\ge0 .075$,
raising its review burden to $32.0\%$---above Opus~5 and GPT-5.5. At the other extreme,
Gemini-3.1-Pro---whose stated confidence takes only five distinct values---can reach no
development-feasible threshold below full review, so it qualifies only by routing
$100\%$ of cases to a human, attaining exactly $a_h$ without deploying. Together these
cases make the central point: optimizing a policy on fixed data is easy; making a
deployment claim that survives on new data is a much stronger bar.

Because Review\% at a single target is a step function of each system's reachable
threshold grid, systems with coarse signals overshoot---$\widehat R-Y$ ranges from
$+0.002$ (GLM-5.2) to $+0.140$ (Gemini-3.1-Pro). The margin is what closes the gap
between a point estimate and a qualified deployment: without it only $46\%$ of policies
qualify across the frontier sweep, against $95\%$ with it. Cross-system review
burden is therefore best compared on the full frontier of
Figure~\ref{fig:clinicare-frontier}, not at a single target.

\begin{table}[h!]
\centering
\small
\begin{tabular}{lccccc}
\toprule
\textbf{System} & \textbf{Review\%} & \textbf{$\widehat R$} & \textbf{95\% LCB} & \textbf{$a_h^{\min}$} & \textbf{Qualified} \\
\midrule
\multicolumn{6}{@{}l}{\texttt{Claude Code}}\\
\quad Opus 5 & 22.4 & 0.839 & 0.807 & 0.690 & \checkmark \\
\quad Sonnet 5 & 29.6 & 0.856 & 0.826 & 0.677 & \checkmark \\
\addlinespace
\multicolumn{6}{@{}l}{\texttt{Codex}}\\
\quad GPT-5.6-Sol & 32.5 & 0.842 & 0.812 & 0.741 & \checkmark \\
\quad GPT-5.6-Luna & 41.6 & 0.796 & 0.764 & 0.890 & \checkmark \\
\quad GPT-5.5 & 21.3 & 0.843 & 0.811 & 0.661 & \checkmark \\
\quad GPT-5.4 & 39.2 & 0.828 & 0.797 & 0.805 & \checkmark \\
\quad GPT-5.4-mini & 49.1 & 0.850 & 0.822 & 0.773 & \checkmark \\
\addlinespace
\multicolumn{6}{@{}l}{\texttt{Gemini CLI}}\\
\quad Gemini-3.6-Flash & 38.4 & 0.850 & 0.821 & 0.742 & \checkmark \\
\quad Gemini-3.5-Flash & 38.7 & 0.847 & 0.818 & 0.751 & \checkmark \\
\quad Gemini-3.1-Pro & 100.0 & 0.900 & 0.900 & 0.760 & \checkmark$^{\dagger}$ \\
\addlinespace
\multicolumn{6}{@{}l}{\texttt{opencode}}\\
\quad DeepSeek-V4-Pro & 44.0 & 0.839 & 0.810 & 0.787 & \checkmark \\
\quad DeepSeek-V4-Flash & 37.1 & 0.832 & 0.802 & 0.787 & \checkmark \\
\quad GLM-5.2 & 10.9 & 0.762 & 0.725 & \textemdash & -- \\
\quad Qwen-3.7-Plus & 61.9 & 0.874 & 0.851 & 0.752 & \checkmark \\
\quad MiniMax-M3 & 33.9 & 0.809 & 0.777 & 0.851 & \checkmark \\
\quad Kimi-K2.7-Code & 67.7 & 0.860 & 0.837 & 0.786 & \checkmark \\
\bottomrule
\end{tabular}
\caption{\textbf{Held-out deployment qualification at $Y=0.76$.} One row per system, $N_{\mathrm{qual}}=375$, review assumption $a_h=0.90$. Thresholds are selected on the development half at $Y+0.05$ and frozen before qualification; the margin absorbs selection bias, without which only 46\% of policies qualify across the frontier sweep, against 95\% with it; at this target, 15 of 16.
 Following Equation~\ref{eq:reliability-lcb-assumed-human}, confidence is attached only to the observed component (exact Clopper--Pearson on accepted cases), none to the assumed $a_h$, so every status is conditional on that assumption. $a_h^{\min}$ (Equation~\ref{eq:break-even-review-assumption}) is the weakest review-success rate under which the policy still qualifies; \textemdash{} marks policies qualifying under no $a_h\in[0,1]$. A system reviewing 100\% of cases attains exactly $a_h$ and qualifies \emph{without deploying}, so systems are compared on Review\%. $^{\dagger}$Qualifies only by routing every case to review, attaining exactly $a_h$; not a deployment. }
\label{tab:clinicare-qualification}
\end{table}

\subsection{Similar Autonomous Performance, Different Deployment Value}
\label{sec:empirical-hero-comparison}

The central \EFE{} hypothesis is that autonomous benchmark performance alone does not
determine deployment value. To test it, we compare two systems with nearly identical
full-autonomy accuracy but different routing quality, with the pair chosen on
$S_{\mathrm{dev}}$ and frozen before we read the held-out results.

A capability leaderboard would rank GPT-5.4 ($a_0=72.8$) and Sonnet~5 ($a_0=72.5$) as
essentially tied---$0.3$ points apart---yet their stated-confidence signals differ in
quality, with Sonnet~5 the stronger router (AUROC $0.681$ vs.\ $0.601$; AURC $0.203$
vs.\ $0.222$). On $S_{\mathrm{qual}}$ at $Y=0.76$ and $a_h=0.90$, both qualify, but at
markedly different oversight cost: Sonnet~5 meets the target while routing only $29.6\%$
of cases to human review ($\widehat R=0.856$, LCB $0.826$), whereas GPT-5.4 requires
$39.2\%$ ($\widehat R=0.828$, LCB $0.797$)---a third more human work for the same
reliability. The gap reflects routing quality, not accuracy: because Sonnet~5's
confidence more sharply separates its own successes from its failures, a less aggressive
threshold already clears the reliability target and leaves more work safely autonomous.

Better routing does not translate into less review monotonically at a single target,
and we report a case that shows why rather than suppress it. Qwen-3.7-Plus and
MiniMax-M3 have identical development accuracy ($a_0=66.4$) and Qwen has the better
signal on both AURC and AUROC, yet Qwen requires $61.9\%$ review at $Y=0.76$ against
MiniMax's $33.9\%$. The cause is overshoot: Qwen's nearest reachable threshold delivers
$\widehat R=0.874$, paying for $11$ points of reliability the target did not
require, so single-target review burden is a noisy summary, and the frontier of
Figure~\ref{fig:clinicare-frontier} is the fairer cross-system comparison.

Two systems can thus post near-identical benchmark scores yet support different
operating policies, because one more effectively identifies the cases on which
autonomous execution will fail. The difference surfaces not as another accuracy number
but as the human oversight needed to reach the same reliability---the
capability-versus-deployability distinction that motivates \EFE{}.
\subsection{Scope and Additional Analyses}
\label{sec:empirical-scope}

This analysis deliberately isolates terminal accept-or-escalate routing so the
relationship among routing quality, reliability, and review burden stays transparent;
it is not a replacement for the full CliniCARE evaluation. A richer deployment could
enlarge the action space---for example, requesting additional evidence before
adjudication---but doing so makes the oversight action depend on the workflow's own
state and can change the subsequent trajectory, moving into the trajectory-dependent
setting of Section~\ref{sec:trajectory-dependent-oversight} rather than the
signal-based policy studied here.

Appendix~\ref{app:clinicare-supplementary} reports the supporting analyses:
sensitivity of the frontier to the assumed review success $a_h$, robustness under the
process-aware success definition
(Equation~\ref{eq:clinicare-fourway-defectfree-success}), and complete per-system
results. Taken together, the case study demonstrates the deployment-level information
\EFE{} adds: the same task evidence used by a conventional benchmark is translated into
a statistically qualified relationship among autonomous performance, routing quality,
human oversight, and operating reliability.

%% file: sections/related.tex
\section{Related Work}
\label{sec:related}

\begin{table}[!htbp]
\centering
\footnotesize
\setlength{\tabcolsep}{3pt}
\renewcommand{\arraystretch}{1.25}
\renewcommand{\tabularxcolumn}[1]{m{#1}}
\begin{tabularx}{\textwidth}{@{} >{\RaggedRight\hyphenpenalty=10000\arraybackslash}m{2.4cm} Y *{6}{c} @{}}
\toprule
\textbf{Work / family}
& \textbf{Primary question}
& \rotatebox[origin=l]{90}{\textbf{Executable task}}
& \rotatebox[origin=l]{90}{\textbf{Interactive}}
& \rotatebox[origin=l]{90}{\textbf{Reliability}}
& \rotatebox[origin=l]{90}{\textbf{Operating cost}}
& \rotatebox[origin=l]{90}{\textbf{Human oversight}}
& \rotatebox[origin=l]{90}{\textbf{Statistical qual.}}
\\
\midrule
\rowcolor{scaleForest!14}
\textbf{\EFE{} (ours)}
& Under what conditions and at what cost can an organization reliably deploy an agent on a specified workflow?
& \yes & \yes & \yes & \yes & \yes & \yes
\\
\midrule
\multicolumn{8}{@{}l}{\textit{(\S\ref{sec:related-capability}) Capability and work-product benchmarks}}\\
\rowcolor{scaleLightGray!55}
Humanity's Last Exam~\citep{hle}
& Does the model know the answer to closed-ended, frontier-difficulty questions?
& \nmark & \nmark & \nmark & \nmark & \nmark & \nmark
\\
Agents' Last Exam~\citep{agentslastexam}
& Can a generalist computer-use agent complete authentic, long-horizon professional work?
& \yes & \nmark & \nmark & \nmark & \nmark & \nmark
\\
\rowcolor{scaleLightGray!55}
GDPval~\citep{gdpval}
& Is the model-generated work product as good as an expert's?
& \yes & \nmark & \nmark & \pmark & \nmark & \nmark
\\
$\tau^2$-bench~\citep{tautwobench}
& Can an agent coordinate with a user and tools to complete a task reliably across trials?
& \yes & \yes & \yes & \nmark & \nmark & \nmark
\\
\rowcolor{scaleLightGray!55}
CRMArena-Pro~\citep{crmarenapro}
& Can an agent perform professional CRM tasks across business functions and interaction types?
& \yes & \pmark & \nmark & \nmark & \nmark & \nmark
\\
AI Agents That Matter~\citep{kapoor2024agentsmatter}
& Do agent benchmarks reward accuracy while ignoring cost?
& \yes & \nmark & \nmark & \yes & \nmark & \pmark
\\
\midrule
\multicolumn{8}{@{}l}{\textit{(\S\ref{sec:related-process-evaluation}) Process, evidence, and trajectory evaluation}}\\
\rowcolor{scaleLightGray!55}
Process / trajectory eval.~\citep{sun2025collab,chen2026mlr}
& Are the intermediate steps, tool use, and evidence correct, not just the final answer?
& \yes & \nmark & \nmark & \nmark & \nmark & \nmark
\\
\midrule
\multicolumn{8}{@{}l}{\textit{(\S\ref{sec:related-selective-prediction}) Selective prediction, abstention, and learning to defer}}\\
\rowcolor{scaleLightGray!55}
Learning to defer~\citep{madras2018predict,mozannar2020consistent}
& Should the model predict this case or defer it to a human?
& \nmark & \nmark & \nmark & \pmark & \yes & \nmark
\\
\midrule
\multicolumn{8}{@{}l}{\textit{(\S\ref{sec:related-routing-signals}) Confidence, uncertainty, and calibration}}\\
\rowcolor{scaleLightGray!55}
Confidence / calibration~\citep{kadavath2022language,kuhn2023semantic}
& How likely is this output correct, and is that confidence calibrated?
& \nmark & \nmark & \pmark & \nmark & \nmark & \nmark
\\
\midrule
\multicolumn{8}{@{}l}{\textit{(\S\ref{sec:related-cost-routing}) Cost-aware routing, cascades, and deferral economics}}\\
\rowcolor{scaleLightGray!55}
Cost-aware routing \& cascades~\citep{frugalgpt2023,ong2025routellm}
& Can we hit a quality bar at lower cost by routing between models?
& \pmark & \nmark & \pmark & \yes & \nmark & \nmark
\\
\midrule
\multicolumn{8}{@{}l}{\textit{(\S\ref{sec:related-oversight-assurance}) Human oversight, scalable oversight, and assurance}}\\
\rowcolor{scaleLightGray!55}
Human oversight \& assurance~\citep{vaccaro2024combinations,mitchell2019modelcards}
& Does human review of AI improve outcomes and provide deployment assurance?
& \nmark & \pmark & \pmark & \nmark & \yes & \pmark
\\
\bottomrule
\end{tabularx}
\caption{
\textbf{\EFE{} as the integration point across the six related-work families of Section~\ref{sec:related}.}
Each row group corresponds to one subsection (\S\ref{sec:related-capability}--\S\ref{sec:related-oversight-assurance}), with a representative work or method, along six axes: whether it runs an agent on \emph{executable} multi-step work; whether it involves an \emph{interactive} user or counterparty; whether it measures \emph{reliability} against a target; whether it accounts for operating \emph{cost}; whether it models \emph{human oversight}; and whether it produces a \emph{statistically qualified} deployment claim on held-out data.
\yes~= addressed, \pmark~= partial or implicit, \nmark~= not addressed. \EFE{} is the only approach that integrates all six into a single deployment-qualification claim.
}
\label{tab:positioning}
\end{table}

\EFE{} is not another benchmark of professional capability.
It adds a deployment-qualification layer on top of such benchmarks:
given evidence of how well an agent performs a workflow, it asks under what
oversight policy, at what cost, and with what statistical support the agent can
be deployed at a target reliability.
This section situates \EFE{} against six bodies of work:
capability benchmarks for knowledge and professional agents
(Section~\ref{sec:related-capability}); outcome, process, and evidence-based evaluation
(Section~\ref{sec:related-process-evaluation});
selective prediction, abstention, and learning to defer
(Section~\ref{sec:related-selective-prediction});
confidence, uncertainty, and calibration as routing signals
(Section~\ref{sec:related-routing-signals}); cost-aware routing, cascades, and deferral economics (Section~\ref{sec:related-cost-routing}); and human oversight, scalable oversight, and deployment assurance (Section~\ref{sec:related-oversight-assurance}); and, finally, the general agent-evaluation frameworks \EFE{} builds on rather than competes with (Section~\ref{sec:related-frameworks}). 
\EFE{} draws on all six but integrates them into a common deployment-qualification framework that evaluates the reliability, oversight burden, and operating cost of the deployed human--AI system.

Table~\ref{tab:positioning} organizes this section. Each row group corresponds to
one of the families surveyed below, and the columns are the deployment dimensions
\EFE{} integrates: executable and interactive task evidence, measured reliability,
operating cost, human oversight, and statistical qualification. Reading across the
\EFE{} row, every dimension is addressed; reading across any other row, the adjacent
literatures each supply some dimensions but none combine them into a
deployment-qualification claim. \EFE{} is the integration point across all six.

\subsection{From Knowledge Exams to Professional-Work Agents}
\label{sec:related-capability}

Evaluation of frontier models has progressed from static knowledge exams to
executable professional work.
Closed-ended exams such as Humanity's Last Exam~\citep{hle} measure expert
knowledge at the frontier of human difficulty, but they score isolated answers
not an agent's ability to carry out a task or to be relied upon in
operation.
Agentic benchmarks instead evaluate whether a system can \emph{do} the work.
Agents' Last Exam~\citep{agentslastexam} targets authentic, long-horizon
computer-use workflows;
GDPval~\citep{gdpval} compares model deliverables against expert work products
and estimates the associated time or cost savings;
xbench~\citep{xbench} tracks profession-aligned productivity and
technology--market fit; and
recent work formalizes how to design and report benchmarks for knowledge
work~\citep{workactivitytax}.

A large surrounding literature exercises the enterprise and tool-use skills
these workflows depend on.
General-assistant and web benchmarks include GAIA~\citep{gaia},
WebArena and VisualWebArena~\citep{webarena,visualwebarena}, and
OSWorld~\citep{osworld};
enterprise task suites include WorkArena~\citep{workarena},
CRMArena and CRMArena-Pro~\citep{crmarena,crmarenapro}, and
Terminal-Bench~\citep{merrill2026terminalbench};
conversational tool-use with a simulated user is studied by
$\tau$-bench and $\tau^2$-bench~\citep{taubench,tautwobench}; and
retrieval- and browsing-heavy work is covered by
BrowseComp~\citep{browsecomp} and deep-research
benchmarks~\citep{deepresearchbench,drbench}. Recent surveys further catalog this rapidly growing space~\citep{yehudai2025agenteval,mohammadi2025agentbenchsurvey}, and enterprise-grade benchmarks now span software engineering~\citep{jimenez2024swebench}, simulated companies~\citep{xu2024agentcompany}, workplace tasks~\citep{styles2024workbench,trivedi2024appworld}, and enterprise data analytics~\citep{lei2024spider2}, building on general agent benchmarks such as AgentBench~\citep{liu2024agentbench}. A parallel line evaluates the safety and trustworthiness of agent behavior beyond task success~\citep{agentharm,stwebagentbench,agentsafetybench}. 

However, these evaluations primarily characterize whether the work gets done and how well, although recent work has begun to expose additional operational dimensions. \citet{kapoor2024agentsmatter} show that accuracy-only agent benchmarks can favor needlessly costly systems and advocate jointly reporting cost and accuracy on held-out data, while $\tau$-bench's \emph{pass\textsuperscript{k}} measures whether an agent succeeds across $k$ independent attempts and thereby captures consistency beyond single-attempt capability~\citep{taubench}. \EFE{} builds on these richer forms of task evidence but addresses a distinct deployment question: which oversight policy meets a prespecified reliability target at minimum operating cost, and does held-out evidence statistically support that operating point? Thus, \EFE{} consumes rather than replaces capability benchmarks, adding a reliability--oversight--cost qualification layer for deployment. 
The first row group of Table~\ref{tab:positioning} places these capability and
work-product benchmarks against \EFE{}'s deployment dimensions.

\subsection{Outcome, Process, and Evidence-Based Evaluation}
\label{sec:related-process-evaluation}


Recent work increasingly recognizes that endpoint task success alone may provide an incomplete account of agent behavior. Process- and trajectory-oriented evaluations therefore examine intermediate execution behavior in addition to final outcomes, including whether agents follow appropriate procedures, use tools correctly, and satisfy task-specific intermediate requirements \cite{sun2025collab,chen2026mlr}. Related work on factuality and grounding evaluates whether generated outputs are supported by available evidence, while recent studies further show that conclusions about hallucination or grounding can depend substantially on the evaluation protocol itself \cite{geigle2024does,janiak2025illusion}. Other approaches use consistency or model-internal signals to detect unsupported generation \cite{manakul2023selfcheckgpt,binkowski2025hallucination,orgad2024llms}, while work on reasoning faithfulness demonstrates that plausible model explanations need not faithfully reflect the processes that produced an answer \cite{lanham2023measuring,turpin2023language}. More broadly, recent reviews of agent evaluation argue that deployment-oriented assessment must extend beyond aggregate task completion to consider execution quality, evidence, safety, and other workflow-specific requirements \cite{kehkashan2026benchmarks}.

\EFE{} builds on these ideas but addresses a distinct deployment-qualification problem. It does not define a universal truthfulness or grounding metric.
These properties enter through the workflow evaluator $g_w$ when they matter for
success, for example, a workflow may require evidence grounding, provenance,
required tool use, or appropriate abstention alongside outcome correctness, and
the success predicate $\phi_w$ specifies which measurements an execution must
satisfy.
Crucially, any process requirement used for qualification must be supported by
observable execution evidence, such as retrieved sources, tool calls, or state
changes, not unrecorded model reasoning.
\EFE{} treats outcome and process evaluation as workflow-specific inputs
to a common reliability-and-oversight methodology, not as universal properties
of an agent.

\subsection{Selective Prediction, Abstention, and Learning to Defer}
\label{sec:related-selective-prediction}

\EFE{}'s terminal accept-or-escalate setting is an instance of selective prediction, in which a system answers on a subset of cases and abstains on the remainder, trading coverage against error among accepted cases. This risk--coverage formulation has a long history: the optimal reject rule dates to \citet{chow1970optimum}, selective classification was formalized in later work~\citep{elyaniv2010selective}, and confidence-based selection was extended to deep networks~\citep{geifman2017selective}. The reject-option and abstention settings are surveyed by \citet{hendrickx2024reject} and, for language models, by \citet{wen2025abstention}, with optimal-strategy characterizations in \citet{franc2023optimal}. A closely related literature studies \emph{learning to defer}, in which a model chooses between predicting and passing a case to a downstream decision-maker~\citep{madras2018predict,mozannar2020consistent}. Subsequent work develops calibrated deferral~\citep{verma2022calibrated}, exact deferral algorithms~\citep{mozannar2023who}, human-complementary predictors~\citep{charusaie2022sample}, and coverage-constrained multi-expert cooperation~\citep{zhang2024coverage}. Distribution-free guarantees for selective decisions and thresholds are surveyed in \citet{angelopoulos2021gentle,campos2024conformalnlp}.

Recent work extends these ideas to language models. AbstentionBench~\citep{abstentionbench} evaluates whether models appropriately abstain on unanswerable questions; conformal methods construct coverage guarantees for LLM predictions, including settings without logit access~\citep{kumar2023conformal,su2024api}; and uncertainty-aware planners request human assistance when uncertainty is high~\citep{ren2023robots}. These approaches inherit the same basic tradeoff between automated coverage and error among accepted cases. Accordingly, the risk--coverage curve and its area (AURC), which we use in Section~\ref{sec:routing-quality} to summarize routing quality, come directly from the selective-prediction literature~\citep{elyaniv2010selective,geifman2017selective}. Likewise, the human-review pathway evaluated by \EFE{} is closely connected to the classical learning-to-defer formulation, in which responsibility is allocated between the model and a downstream decision-maker~\citep{madras2018predict,mozannar2020consistent}.

\EFE{} builds on selective autonomy but changes what is optimized.
An abstention or deferral rate alone does not determine whether a deployment is
acceptable, because it says nothing about what happens to deferred cases.
\EFE{} instead models the outcome of the declared human-review path and evaluates
the reliability of the human--AI system,
\begin{equation}
R_{m,w}(\pi)=\mathbb{E}[u^\pi],
\end{equation}
together with its expected operating cost $C_{m,w}(\pi)$, and then selects the
oversight policy that meets a reliability target at minimum cost. This formulation is related to work on human-complementary prediction and coverage-constrained cooperation~\citep{charusaie2022sample,zhang2024coverage}, but READY's objective is deployment qualification under a workflow-level reliability requirement rather than optimization of abstention or coverage alone.

This has therefore led to three distinctions. First, human review is part of the evaluated operating policy, not merely an
abstention.
Second, \EFE{} separates policy selection from held-out statistical qualification,
so the reported operating point is supported by independent evidence.
Third, \EFE{} distinguishes \emph{task-level abstention}---a valid workflow
outcome, such as a correct indeterminate verdict in CliniCARE-Bench---from
\emph{deployment-level escalation}, the separate decision to route a completed
result to human review; Appendix~\ref{app:clinicare-supplementary} develops
this distinction. Thus, selective prediction and learning-to-defer methods provide the conceptual foundation for selective autonomy, while READY uses that foundation to qualify an explicit human--AI deployment operating point in terms of reliability, oversight, and cost.

\subsection{Confidence, Uncertainty, and Calibration as Routing Signals}
\label{sec:related-routing-signals}

Terminal oversight requires an observable signal $s_i$ that ranks completed
executions by their likelihood of autonomous success.
Many candidate signals exist.
Models can self-assess correctness through elicited probabilities such as
\emph{P(True)} and \emph{P(IK)}~\citep{kadavath2022language} or token-level
self-evaluation~\citep{ren2023selfeval}.
Verbalized confidence~\citep{lin2022teaching,tian2023just,ji2025calibrating,liu2025metafaith}, relative confidence
preferences~\citep{shrivastava2025relative}, semantic
uncertainty~\citep{kuhn2023semantic,ji2025calibrating}, confidence-weighted
self-consistency~\citep{taubenfeld2025confidence}, input-clarification
ensembles~\citep{hou2024decomposing}, functional
uncertainty~\citep{niu2024functional}, and logit-based
methods~\citep{ma2025inferring} offer further ways to estimate uncertainty.

\EFE{} does not prescribe a signal; it evaluates the routing behavior induced by
whatever signal the deployment policy can observe.
For threshold routing, the score need not be a calibrated probability---it need
only order cases so that more selective acceptance preferentially removes
higher-risk executions (Section~\ref{sec:routing-quality}).
Calibration~\citep{guo2017calibration,xie2024adaptive,li2024graph}, its
behavior in language
models~\citep{braverman2020calibration,jiang2021know,chen2023close,tian2023just},
and training-time methods for improving
it~\citep{kapoor2024calibration,stengeleskin2024lacie,zhao2022slic,zhao2023slichf,shi2024understanding}
remain relevant when the numerical value of the signal is interpreted
probabilistically, but calibration is not a prerequisite for qualification.
Because signals that require repeated sampling or extra inference improve
discrimination at a cost, \EFE{} folds that cost into the operating-cost quantity
$C_{m,w}(\pi)$ instead of treating it as external to the comparison.

Surveys of confidence estimation and uncertainty quantification for language models organize the space of candidate signals~\citep{geng2024confidence,shorinwa2024uqsurvey}, as do hallucination surveys~\citep{ji2023hallucination,huang2023hallucinationllm}. Representative signals include semantic entropy~\citep{farquhar2024semanticentropy}, black-box confidence elicitation~\citep{xiong2024canllms}, internal-state detectors~\citep{chen2024inside}, and meaning-aware scores~\citep{duan2024sar,bakman2024mars}, with large-scale comparisons of which signals are actually discriminative~\citep{vashurin2025lmpolygraph}. Selective generation~\citep{ren2023selectivegen} and conformal generation~\citep{quach2023conformal} apply such signals to the accept-or-defer decision directly, and when a signal is interpreted probabilistically its calibration must be measured carefully~\citep{naeini2015bbq,nixon2019adaptivece}.

\subsection{Cost-Aware Routing, Cascades, and Deferral Economics}
\label{sec:related-cost-routing}

\EFE{} selects the oversight policy that meets a reliability target at minimum
operating cost, which connects it to a large literature on cost-aware inference
for large language models. Model \emph{cascades} invoke a cheap model first and
escalate to a more expensive one when a confidence signal is low~\citep{farinhas2025translate,pmlr-v267-dekoninck25a,shen2025sater}. FrugalGPT
learns such a cascade to hit a target accuracy at reduced
cost~\citep{frugalgpt2023}, AutoMix escalates under noisy self-verification via
a meta-router~\citep{automix2023}, mixture-of-thought cascades defer only hard
reasoning cases~\citep{yue2024cascades}, and EcoAssistant extends the pattern to
code-executing agents~\citep{ecoassistant2023}. A parallel line \emph{routes}
each query to the cheapest adequate model: RouteLLM learns a router from
preference data~\citep{ong2025routellm}, Hybrid LLM routes by predicted
difficulty~\citep{hybridllm2024}, and Zooter routes by predicted
expertise~\citep{zooter2023}, with RouterBench providing a standard cost--quality
evaluation~\citep{routerbench2024}. Recent surveys formalize routing and
cascading as a shared performance--cost optimization
problem~\citep{routingsurvey2025}, a pattern that descends from classical
test-time cost-sensitive cascades in detection~\citep{violajones2001}.

\EFE{} shares the broader goal of reducing operating cost subject to an acceptable performance requirement, but differs in three deployment-relevant respects. First, the deferral
target is a \emph{human reviewer}, not a stronger model, so the escalation
cost is human-review effort and the reliability of the deployed system depends on
the review path, not on a larger model's accuracy. Second, \EFE{} replaces an
average cost--quality operating point with a \emph{statistically qualified}
reliability floor established on held-out data, not a heuristic threshold
or an average win rate. Third, the quality bar is a workflow-specific success
predicate that may include process and policy requirements, not final-answer
correctness alone. Cascade and routing methods are complementary:
their confidence-based escalation logic can supply the routing signal $s_i$,
while \EFE{} supplies the qualification procedure that turns it into a deployment
guarantee.

\subsection{Human Oversight, Scalable Oversight, and Deployment Assurance}
\label{sec:related-oversight-assurance}

\EFE{} evaluates a human--AI system under an oversight policy, connecting most directly to the 
work on human oversight of AI and assurance for deployment. The notion that
supervision is a costly, budgeted resource originates in the scalable-oversight
literature~\citep{amodei2016concrete}; subsequent work measures whether a human
assisted by an unreliable model outperforms either
alone~\citep{bowman2022measuring}, studies oversight when the supervised system
is stronger than its overseer~\citep{burns2023weak}, and proposes debate and
amplification as mechanisms~\citep{irving2018debate,christiano2018amplification,lang2025debate}.
More recent work emphasizes that effective oversight requires meaningful human agency rather than nominal human presence~\citep{zhu2026designing,tsamados2025human}, while empirical studies show that human--AI teaming does not reliably outperform the stronger individual component and depends on factors such as interaction design and reviewer expertise~\citep{tang2026dark,liu2025human}. Particularly, a meta-analysis of 106 studies finds human--AI
combinations on average underperform the better of human or AI
alone~\citep{vaccaro2024combinations}, complementary performance requires more
than model explanations~\citep{bansal2021whole}, and overreliance is a central
failure mode of human review~\citep{bucinca2021trust,lai2021science}. Mandated
human oversight in particular can provide false assurance when reviewers cannot
perform the checking function~\citep{green2022flaws}. Finally, assurance and
auditing frameworks---model cards~\citep{mitchell2019modelcards}, internal
algorithmic auditing~\citep{raji2020closing}, and socio-technical safety
evaluation~\citep{weidinger2023sociotechnical}---establish reporting evaluated
performance under intended deployment conditions.

\EFE{} responds to these findings directly. Rather than assuming human review
improves outcomes, it treats the review path's success rate as a measured
quantity or an explicit deployment assumption, and qualifies the deployed system
on held-out evidence. \EFE{} consequently complements model cards and audit frameworks by reporting a statistically supported deployment profile--the reliability, oversight burden, and operating cost of a specific human--AI configuration-- rather than certifying the agent in isolation

\subsection{General Agent-Evaluation Frameworks}
\label{sec:related-frameworks}

A final, infrastructure-level body of work provides the execution and harness
frameworks on which agent evaluations are built, as distinct from the six
methodological literatures above. Inspect AI~\citep{inspect_ai} standardizes task
specification, agent and tool integration, sandboxed execution, scoring, and evaluation
logging; Harbor~\citep{harbor} packages agent evaluations as containerized tasks with
verification scripts; and AgentBeats~\citep{agentbeats} provides a platform for running and
comparing agent evaluations under a common protocol. These frameworks answer \emph{how} an
evaluation is executed and recorded. \EFE{} is deliberately not a competing harness: its
reference implementation runs on Inspect and can consume Harbor-packaged tasks through the
\texttt{inspect-harbor} adapter. \EFE{} adds the orthogonal layer of \emph{what
deployment operating point} the recorded evidence supports, selecting an oversight policy
and statistically qualifying a reliability--oversight--cost profile on held-out data. In
the terms of Table~\ref{tab:positioning}, these frameworks supply the executable-task
substrate but none of the reliability, cost, oversight, or statistical-qualification
dimensions; \EFE{} is the qualification layer that sits above whichever framework produced
the evidence.

%% file: sections/limits.tex
\section{Scope, Limitations, and Future Work}
\label{sec:limitations}

\EFE{} is intended as a framework for evaluating and statistically qualifying
AI-agent deployment under explicit workflow, oversight, reliability, and cost
assumptions.
It does not attempt to provide a universal certification of an agent or to
capture every organizational consideration involved in production deployment.

\paragraph{Empirical scope.}
The empirical evaluation in this paper focuses on the trajectory-invariant,
terminal accept-or-escalate setting using CliniCARE-Bench.
This setting provides a transparent test of the central \EFE{} methodology:
routing agent outputs between autonomous acceptance and human review to meet
a target reliability at minimum oversight cost.
Section~\ref{sec:trajectory-dependent-oversight} extends the formulation to
trajectory-dependent policies in which review, clarification, correction, or
other intervention changes subsequent execution.
Evaluating such policies requires policy-in-the-loop execution or simulation;
large-scale empirical validation of these settings is left to future work.

\paragraph{Human-oversight assumptions.}
\EFE{} distinguishes measured evaluation evidence from assumptions about the
deployment environment.
In particular, when the declared human-review path is not executed directly,
its success rate, latency, and cost must be supplied as deployment assumptions.
The deployment profile is conditional on those values.
Future evaluations should increasingly measure reviewer effectiveness,
review time, intervention type, and recovery cost directly instead of treating them as fixed parameters.

\paragraph{Population and distribution validity.}
A qualification claim applies to the workflow population represented by the
evaluation and to the versioned configuration under which the evidence was
collected.
It does not guarantee the same reliability under arbitrary distribution
shift, changes in workflow composition, environment changes, or materially
different agent behavior. Continuous post-deployment monitoring is therefore important for detecting
performance degradation and determining when requalification is needed, but
designing such monitoring and requalification triggers is beyond the scope of this work.

\paragraph{Risk and deployment context.}
The primary \EFE{} objective represents workflow success through a
workflow-specific binary event and optimizes expected operating cost
subject to a reliability target.
For workflows in which failures differ substantially in consequence, the
framework permits additional risk constraints, including tail-risk measures.\EFE{} does not replace enterprise security, privacy, legal,
regulatory, or organizational-governance processes.
These considerations may define workflow constraints or evaluation criteria,
but they remain outside the core deployment-qualification methodology.

Future work should extend \EFE{} along three directions:
empirical evaluation of trajectory-dependent and multi-step oversight
policies, direct measurement of human-review effectiveness and economics, and
broader validation across enterprise workflows and domains.
The open interface described in Section~\ref{sec:platform} supports these extensions without changing the underlying qualification objective.

%% file: sections/conclusion.tex
\section{Conclusion}
\label{sec:conclusion}

Professional-work benchmarks measure how well an AI agent can perform a task
autonomously.
Enterprise deployment requires a different determination: whether the agent
can be deployed at a required level of reliability, with an acceptable amount
of human oversight and operating cost.

\efename{} (\EFE{}) formulates this as a deployment-qualification
problem.
For a given workflow, \EFE{} evaluates agent execution, selects an oversight
policy that satisfies a target reliability at minimum operating cost, and
statistically qualifies that operating point on held-out evidence. A deployment profile is not an intrinsic score of the
agent. It is a conditional statement about how that agent can be deployed under
a particular workflow, oversight model, and set of operating assumptions.

\EFE{} separates workflow-specific definitions of successful work from a common
qualification methodology and provides an open runtime testbed for executing
and extending such evaluations.
Trajectory-invariant policies can be evaluated efficiently from saved
execution evidence, while trajectory-dependent policies use the same
optimization objective but require policy-in-the-loop execution or simulation.
CliniCARE-Bench provides an end-to-end instantiation of this framework and
illustrates how autonomous task performance, routing quality, human oversight,
and deployment reliability can be analyzed together.

As agent capabilities improve, the relevant question is not simply whether
benchmark scores rise, but how much human oversight remains necessary to meet
a specified reliability requirement.
\EFE{} provides a framework for measuring that transition and for comparing
systems by the cost of reliable deployment, not autonomous performance alone.

\section*{Author Contributions}
Y.~Xue conceived the project, defined the vision and direction, led the manuscript effort, and supervised the work.
V.~Chatrath led benchmark execution, including the Contributor network and experiments, and served as primary manuscript lead.
B.~Zhu and J.~Fan served as technical and manuscript leads and were primary developers.
G.~Pu was a primary developer.
S.~D.~Tiwari, S.~Dan, and R.~Young developed the benchmark pipeline.
A.~Shanker contributed the mathematical formulation.
Y.~Li, Y.~Yao, and M.~Yang contributed to manuscript iteration.
D.~Y.~Zhang, Y.~He, Y.~Liu, C.~Wang, and Z.~Yin served as research advisors.
All authors contributed to the writing, reviewed and approved the final manuscript.

%% file: sections/appendix.tex
\input{sections/app-metrics}
\input{sections/app-results}

%% file: sections/app-metrics.tex
\section{Formal definitions}
\subsection{Threshold Routing and Monotonicity of the Routing Signal}
\label{app:routing-monotonicity}

Consider the trajectory-invariant terminal-oversight setting.
For task instance $i$, autonomous execution produces an observable routing
signal $s_i \in \mathbb{R}$ and an evaluation outcome
$u_i \in \{0,1\}$, where $u_i=1$ denotes successful autonomous execution.
The outcome $u_i$ is available to the evaluator during policy development
and qualification, but is generally not known when the routing decision is
made.

A threshold policy $\pi_\tau$ accepts the agent output autonomously when
$s_i \geq \tau$ and routes it to human review otherwise:

\begin{equation}
\pi_\tau(s_i)
=
\begin{cases}
\mathrm{accept}, & s_i \geq \tau,\\
\mathrm{review}, & s_i < \tau.
\end{cases}
\label{eq:threshold-routing-policy}
\end{equation}

For threshold $\tau$, define the autonomous coverage

\begin{equation}
c(\tau)
=
\Pr(s \geq \tau),
\label{eq:appendix-routing-coverage}
\end{equation}

and the success probability among autonomously accepted cases,

\begin{equation}
a(\tau)
=
\Pr(u=1 \mid s\geq\tau).
\label{eq:appendix-selective-success}
\end{equation}

We call $s$ a \emph{monotone routing signal} if increasing the threshold does
not decrease the success probability of the retained autonomous cases:

\begin{equation}
\tau_1 < \tau_2
\quad\Longrightarrow\quad
a(\tau_1) \leq a(\tau_2),
\label{eq:routing-monotonicity}
\end{equation}

for thresholds with nonzero autonomous coverage.
Equivalently, increasingly selective autonomous acceptance should
preferentially remove higher-risk cases.

A stronger sufficient condition is pointwise monotonicity,

\begin{equation}
s_1 < s_2
\quad\Longrightarrow\quad
\Pr(u=1\mid s=s_1)
\leq
\Pr(u=1\mid s=s_2),
\label{eq:pointwise-routing-monotonicity}
\end{equation}

but \EFE{} does not require this stronger condition.
In finite evaluation samples, local departures from monotonicity may occur;
what matters operationally is whether the signal induces a sufficiently
ordered risk--coverage relationship to support reliable selective routing.

Importantly, monotonicity is distinct from calibration.
A routing signal may provide a useful ordering of cases even when
$s_i$ is not numerically equal to the probability of successful execution.
Calibration is required only when the numerical value of the signal is
interpreted probabilistically; threshold-based routing principally requires
discrimination and ordering.

\subsection{Tail-Risk Estimation for Risk-Sensitive Qualification}
\label{app:tail-risk-qualification}

The main text treats workflow-specific risk as an optional extension to the
common \EFE{} reliability constraint.
This appendix gives one concrete implementation using conditional
value-at-risk (CVaR).
The construction is not required for workflows in which binary deployment
success adequately captures the consequences of failure.

Let

\begin{equation}
L_i^\pi
=
L_w(Z_i^\pi)
\geq 0
\end{equation}

denote the workflow-specific loss incurred on qualification instance $i$
under a fixed oversight policy $\pi$.
The loss function is defined as part of the workflow specification and may
encode, for example, the severity of a failed action, downstream recovery
cost, or another consequence not represented adequately by the binary
success indicator $u_i^\pi$.

For confidence level $\alpha\in(0,1)$, the value-at-risk is

\begin{equation}
\operatorname{VaR}_{\alpha}(L^\pi)
=
\inf
\left\{
\ell :
\Pr(L^\pi \leq \ell)\geq\alpha
\right\}.
\label{eq:var-definition}
\end{equation}

CVaR measures the expected loss in the upper tail beyond this quantile.
For estimation, we use the equivalent optimization representation

\begin{equation}
\operatorname{CVaR}_{\alpha}(L^\pi)
=
\min_{\eta\in\mathbb{R}}
\left[
\eta
+
\frac{1}{1-\alpha}
\mathbb{E}
\left[
(L^\pi-\eta)_+
\right]
\right],
\label{eq:cvar-definition}
\end{equation}

where $(x)_+=\max(x,0)$.
This form remains well defined for discrete empirical loss distributions and
avoids requiring a separate estimator of the tail quantile.

\paragraph{Held-out estimator.}
After policy selection, the policy $\widehat{\pi}$ and the risk specification
$(L_w,\alpha,B_w)$ are frozen before evaluation on the qualification set.
Given losses

\begin{equation}
L_1^{\widehat{\pi}},
\ldots,
L_N^{\widehat{\pi}},
\end{equation}

the empirical CVaR estimator is

\begin{equation}
\widehat{\operatorname{CVaR}}_{\alpha}
=
\min_{\eta\in\mathbb{R}}
\left[
\eta
+
\frac{1}{(1-\alpha)N}
\sum_{i=1}^{N}
\left(
L_i^{\widehat{\pi}}-\eta
\right)_+
\right].
\label{eq:empirical-cvar}
\end{equation}

For trajectory-invariant policies, the losses may be computed by replaying
the frozen policy over saved qualification trajectories.
For trajectory-dependent policies, the losses must be obtained from
qualification executions or simulations performed under
$\widehat{\pi}$ itself.

\paragraph{Statistical qualification.}
A point estimate of CVaR is not sufficient to establish a deployment-risk
constraint.
\EFE{ therefore constructs a one-sided upper confidence bound

\begin{equation}
\operatorname{CVaR}^{\,1-\delta}_{\mathrm{UCB}}
\left(
\widehat{\pi}
\right)
\end{equation}

at confidence level $1-\delta$ and qualifies the policy with respect to the
tail-risk constraint only if

\begin{equation}
\operatorname{CVaR}^{\,1-\delta}_{\mathrm{UCB}}
\left(
\widehat{\pi}
\right)
\leq B_w.
\label{eq:cvar-qualification}
\end{equation}

In the reference procedure, this upper bound is obtained by bootstrap
resampling the held-out qualification set.
For each bootstrap replicate $b=1,\ldots,B$, \EFE{} resamples qualification
instances, recomputes
$\widehat{\operatorname{CVaR}}_{\alpha}^{(b)}$, and uses the appropriate
upper percentile of the resulting bootstrap distribution as the one-sided
confidence bound.

The bootstrap unit must match the unit of statistical generalization.
If multiple stochastic runs are collected for the same task instance, all
runs from that instance are resampled together.
Likewise, if observations are clustered by patient, customer, account, or
another natural unit, resampling is performed at that cluster level rather
than treating individual observations as independent.

\paragraph{Uncertain downstream outcomes.}
If $L_i^\pi$ depends on an outcome that is not directly observed in the
evaluation, for example, the success or cost of a hypothetical human-review
path, the corresponding uncertainty must be propagated into the risk
estimate.
\EFE{} may do this by jointly sampling the uncertain review parameters and the
qualification instances in each bootstrap replicate.
A risk quantity computed by substituting a single assumed review-success
rate or recovery cost should therefore be reported as conditional on that
assumption rather than as an empirically qualified tail-risk estimate.

\paragraph{Pre-specification and tail support.}
The loss function $L_w$, tail level $\alpha$, risk tolerance $B_w$, and
confidence level $1-\delta$ must be specified before examining the held-out
qualification results.
They must not be tuned to make a candidate policy pass qualification.

Tail-risk estimates can be statistically weak when the qualification set
contains few observations in the relevant tail.
For example, an $\alpha=0.99$ CVaR estimate is informed primarily by roughly
the worst one percent of the sampled outcomes.
\EFE{ therefore reports the effective number of observations contributing to
the estimated tail together with the point estimate and confidence bound.
If the available qualification sample does not support a sufficiently
precise upper bound, the appropriate result is
\emph{insufficient evidence for risk qualification}, rather than a relaxed
risk threshold.

%% file: sections/app-results.tex
\section{Supplementary Analyses for CliniCARE-Bench}
\label{app:clinicare-supplementary}

This appendix supplements the four-way deployment-qualification analysis of
Section~\ref{sec:empirical} with the held-out lower-bound computation, sensitivity checks, and complete per-system results.
All policies follow the global-routing definition and margin-adjusted
threshold selection of Section~\ref{sec:empirical-policies}; all thresholds
are selected on the development partition and frozen before evaluation on
the held-out qualification partition.

\subsection{Reproducing the Held-Out Lower Bound}
\label{app:clinicare-lcb-procedure}

The $95\%$ LCB column of Table~\ref{tab:clinicare-qualification} instantiates
Equation~\ref{eq:reliability-lcb-assumed-human}, attaching confidence only to the
observed accepted component. Fix the threshold $\widehat\tau$ selected on
$S_{\mathrm{dev}}$ at $Y+\delta$ ($\delta=0.05$;
Equation~\ref{eq:threshold-optimization-with-margin}) and let $N_{\mathrm{qual}}=375$.
\begin{enumerate}
  \item \textbf{Route.} For each qualification case set
  $d_i=\mathbf{1}\{s_i\ge\widehat\tau\}$ (Equation~\ref{eq:qualification-routing}). Let
  $n_{\mathrm{acc}}=\sum_i d_i$ be the number of accepted cases and
  $k_{\mathrm{acc}}=\sum_i d_i\,u_i$ the number of those that are autonomously
  successful, so the reviewed fraction is
  $(N_{\mathrm{qual}}-n_{\mathrm{acc}})/N_{\mathrm{qual}}$ and
  $\widehat p_{\mathrm{acc}}=k_{\mathrm{acc}}/n_{\mathrm{acc}}$.
  \item \textbf{Point estimate.} Combine the observed accepted rate with the assumed
  review success via Equation~\ref{eq:qualification-reliability-decomposition-second}:
  $\widehat R = \tfrac{n_{\mathrm{acc}}}{N_{\mathrm{qual}}}\widehat p_{\mathrm{acc}}
  + \tfrac{N_{\mathrm{qual}}-n_{\mathrm{acc}}}{N_{\mathrm{qual}}}\,a_h$.
  \item \textbf{Lower bound.} Compute the one-sided $95\%$ exact Clopper--Pearson lower
  bound $\underline p_{\mathrm{acc}}$ for the accepted-case success probability from
  $(k_{\mathrm{acc}},n_{\mathrm{acc}})$, i.e.\ the $\alpha=0.05$ quantile of the
  $\mathrm{Beta}\!\left(k_{\mathrm{acc}},\,n_{\mathrm{acc}}-k_{\mathrm{acc}}+1\right)$
  distribution (with $\underline p_{\mathrm{acc}}=0$ if $k_{\mathrm{acc}}=0$). No
  confidence is attached to the assumed term, so
  \[
  \text{95\% LCB}
  = \frac{n_{\mathrm{acc}}}{N_{\mathrm{qual}}}\,\underline p_{\mathrm{acc}}
  + \frac{N_{\mathrm{qual}}-n_{\mathrm{acc}}}{N_{\mathrm{qual}}}\,a_h .
  \]
  \item \textbf{Qualify.} The policy qualifies iff $\text{95\% LCB}\ge Y$.
  \item \textbf{Break-even.} $a_h^{\min}$
  (Equation~\ref{eq:break-even-review-assumption}) is the smallest $a_h\in[0,1]$ for
  which the LCB expression reaches $Y$; because the LCB is nondecreasing in $a_h$ this
  is a one-dimensional solve, and \textemdash{} marks policies for which no
  $a_h\in[0,1]$ suffices.
\end{enumerate}

\subsection{Sensitivity to the Human-Review Assumption}
\label{app:clinicare-ah-sensitivity}

The main analysis treats human-review success $a_h$ as a declared deployment
assumption (Equation~\ref{eq:empirical-human-review-assumption}).
Here we recompute the reliability--oversight frontier and the qualification
outcomes of Section~\ref{sec:empirical-reliability-oversight} under
alternative values of $a_h$, reporting how the required review burden and
qualified/not-qualified status change with the assumed quality of the human
review path.
We repeat the entire dev-select/freeze/qualify procedure at
$a_h\in\{0.80,0.85,0.90,0.95,1.00\}$ over the same $31$-target sweep, giving
$31\times16=496$ policies per assumption.
Nothing about the execution evidence changes; only the declared assumption does.

Table~\ref{tab:app-ah-sensitivity} reports the per-system consequences at the
representative target $Y=0.76$.
The aggregate picture over the full sweep is that the review assumption, not
agent capability, governs how much of the reliability range is reachable at all:

\begin{center}
\small
\begin{tabular}{lccc}
\toprule
$a_h$ & Qualified & Mean Review\% & Highest $Y$ reached by a \emph{deploying} policy \\
\midrule
$0.80$ & $318/496$ & $76.3$ & $0.775$ \\
$0.85$ & $472/496$ & $65.1$ & $0.805$ \\
$0.90$ & $469/496$ & $53.0$ & $\geq 0.850$ \\
$0.95$ & $467/496$ & $44.6$ & $\geq 0.850$ \\
$1.00$ & $463/496$ & $38.9$ & $\geq 0.850$ \\
\bottomrule
\end{tabular}
\end{center}

Weakening the assumed review path from $a_h=0.90$ to $0.80$ raises mean review
burden by $23$ percentage points and caps the highest reliability any deploying
policy attains at $0.775$; at that assumption $311$ of $496$ policies route
every case to review, qualifying only by not deploying.
The last column is censored from above at $a_h\geq0.90$ because the sweep ends
at $Y=0.85$, so those entries are lower bounds rather than ceilings.

Two features of the table deserve comment because they are properties of the
qualification procedure rather than of the systems.
First, the number of qualifying policies is \emph{not} monotone in $a_h$
($472$ at $a_h=0.85$ against $469$ at $a_h=0.90$).
A more generous review assumption relaxes the development-set constraint, which
lets the optimization select a lower threshold and accept more cases; the
resulting accepted set is larger but less reliable, and can fail the
lower-confidence-bound test on the held-out partition that the more
conservative policy passed.
GPT-5.6-Luna at $Y=0.76$ is the clean instance: it qualifies at $a_h=0.80$ and
at $a_h=0.90$, but not at $a_h=0.85$.
This is the expected behaviour of selecting on one partition and testing on
another, and it is a reason to read the frontier rather than any single cell.
Second, review burden is weakly decreasing in $a_h$ for every system, but in
steps, because each system can only move between the thresholds its stated
confidence actually reaches.
Gemini-3.1-Pro is unchanged at $100\%$ review for every $a_h\le0.95$ and only
drops to $33.6\%$ at $a_h=1.00$, which reflects the granularity limit of
Section~\ref{sec:empirical-routing}, not a response to the assumption.

\begin{table}[h!]
\centering
\small
\begin{tabular}{lcccccccccc}
\toprule
& \multicolumn{2}{c}{$a_h=0.80$} & \multicolumn{2}{c}{$a_h=0.85$}
& \multicolumn{2}{c}{$a_h=0.90$} & \multicolumn{2}{c}{$a_h=0.95$}
& \multicolumn{2}{c}{$a_h=1.00$} \\
\cmidrule(lr){2-3}\cmidrule(lr){4-5}\cmidrule(lr){6-7}\cmidrule(lr){8-9}\cmidrule(lr){10-11}
\textbf{System} & Rev\% & Q & Rev\% & Q & Rev\% & Q & Rev\% & Q & Rev\% & Q \\
\midrule
\multicolumn{11}{@{}l}{\texttt{Claude Code}}\\
\quad Opus 5 & 22.4 & \checkmark & 22.4 & \checkmark & 22.4 & \checkmark & 22.4 & \checkmark & 10.4 & \checkmark \\
\quad Sonnet 5 & 41.6 & \checkmark & 29.6 & \checkmark & 29.6 & \checkmark & 25.6 & \checkmark & 25.6 & \checkmark \\
\addlinespace
\multicolumn{11}{@{}l}{\texttt{Codex}}\\
\quad GPT-5.6-Sol & 44.5 & \checkmark & 35.5 & \checkmark & 32.5 & \checkmark & 32.5 & \checkmark & 32.5 & \checkmark \\
\quad GPT-5.6-Luna & 56.0 & \checkmark & 41.6 & -- & 41.6 & \checkmark & 41.6 & \checkmark & 41.6 & \checkmark \\
\quad GPT-5.5 & 35.7 & \checkmark & 35.7 & \checkmark & 21.3 & \checkmark & 15.5 & \checkmark & 15.5 & \checkmark \\
\quad GPT-5.4 & 68.8 & \checkmark & 49.9 & \checkmark & 39.2 & \checkmark & 29.3 & \checkmark & 27.2 & \checkmark \\
\quad GPT-5.4-mini & 100.0 & \checkmark & 67.2 & \checkmark & 49.1 & \checkmark & 39.5 & \checkmark & 36.3 & \checkmark \\
\addlinespace
\multicolumn{11}{@{}l}{\texttt{Gemini CLI}}\\
\quad Gemini-3.6-Flash & 100.0 & \checkmark & 38.4 & \checkmark & 38.4 & \checkmark & 38.4 & \checkmark & 38.4 & \checkmark \\
\quad Gemini-3.5-Flash & 100.0 & \checkmark & 100.0 & \checkmark & 38.7 & \checkmark & 38.7 & \checkmark & 38.7 & \checkmark \\
\quad Gemini-3.1-Pro & 100.0 & \checkmark & 100.0 & \checkmark & 100.0 & \checkmark & 100.0 & \checkmark & 33.6 & \checkmark \\
\addlinespace
\multicolumn{11}{@{}l}{\texttt{opencode}}\\
\quad DeepSeek-V4-Pro & 100.0 & \checkmark & 44.0 & \checkmark & 44.0 & \checkmark & 44.0 & \checkmark & 44.0 & \checkmark \\
\quad DeepSeek-V4-Flash & 100.0 & \checkmark & 84.5 & \checkmark & 37.1 & \checkmark & 37.1 & \checkmark & 37.1 & \checkmark \\
\quad GLM-5.2 & 37.6 & \checkmark & 10.9 & -- & 10.9 & -- & 10.9 & -- & 10.9 & -- \\
\quad Qwen-3.7-Plus & 100.0 & \checkmark & 61.9 & \checkmark & 61.9 & \checkmark & 61.9 & \checkmark & 31.5 & \checkmark \\
\quad MiniMax-M3 & 100.0 & \checkmark & 42.7 & \checkmark & 33.9 & \checkmark & 33.9 & \checkmark & 33.9 & \checkmark \\
\quad Kimi-K2.7-Code & 100.0 & \checkmark & 67.7 & \checkmark & 67.7 & \checkmark & 33.9 & \checkmark & 29.1 & \checkmark \\
\bottomrule
\end{tabular}
\caption{\textbf{Held-out qualification at $Y=0.76$ under alternative
human-review assumptions.} $N_{\mathrm{qual}}=375$; thresholds reselected on
$S_{\mathrm{dev}}$ at $Y+0.05$ for each $a_h$ and frozen. Rev\% is the fraction
of qualification cases routed to human review; Q marks whether the $95\%$ LCB
reaches $Y$. A system at Rev\% $=100$ attains exactly $a_h$ and qualifies
without deploying whenever $a_h\geq Y$.}
\label{tab:app-ah-sensitivity}
\end{table}

\subsection{Process-Aware Success Definition}
\label{app:clinicare-defect-free}

The main analysis scores an autonomous execution as successful when the
four-way verdict matches the reference adjudication
(Equation~\ref{eq:clinicare-fourway-success}).
Here we repeat the principal comparisons using the process-aware variant
$u_{i,\mathrm{df}}$
(Equation~\ref{eq:clinicare-fourway-defectfree-success}), which additionally
requires the absence of a disqualifying process defect, to test sensitivity
of the routing-quality and frontier conclusions to the deployment-success
definition.
CliniCARE records the defect as a single per-run indicator, and it is common:
$21.9\%$ of the $12{,}000$ runs carry one, ranging from $12.7\%$ (GPT-5.5) to
$30.0\%$ (Kimi-K2.7-Code).
Because a defect can co-occur with a correct verdict, substituting
$u_{i,\mathrm{df}}$ for $u_i$ lowers autonomous success by $4.5$ to $13.1$
percentage points depending on the system.

The conclusions of Section~\ref{sec:empirical-routing} survive the substitution;
the frontier of Section~\ref{sec:empirical-reliability-oversight} does not.
Routing quality is measuring the same underlying property under either
predicate: the Spearman correlation between AUROC under $u$ and under
$u_{\mathrm{df}}$ across the $16$ systems is $0.87$, and per-system AUROC moves
by at most $0.032$ (Table~\ref{tab:app-defectfree}).
Qwen-3.7-Plus remains among the best-ranking signals and Gemini-3.1-Pro the
worst.
The deployment consequences, by contrast, shift substantially.
Mean review burden across the sweep rises from $53.0\%$ to $69.8\%$, and at
$Y=0.76$ both Gemini Flash systems move from roughly $38\%$ review to $100\%$:
under a process-aware success definition they have no reachable operating point
that deploys at all.
This strengthens rather than weakens the comparison of
Section~\ref{sec:empirical-hero-comparison}, since the pair contrasted there
collapses to two undeployable systems once process defects count against
success.

Two further observations.
The count of qualifying policies \emph{rises} under $u_{\mathrm{df}}$ ($488$ of
$496$ against $469$), for the same selection reason discussed in
Appendix~\ref{app:clinicare-ah-sensitivity}: a harder success predicate forces
the development-set optimization to far more conservative thresholds, and those
policies clear the held-out bound more comfortably.
GLM-5.2 is the visible case, moving from not qualifying at $10.9\%$ review under
$u$ to qualifying at $44.8\%$ review under $u_{\mathrm{df}}$.
A system's qualified status is therefore not a monotone function of how
demanding the success predicate is, and reporting review burden alongside it
remains necessary.

\begin{table}[h!]
\centering
\small
\begin{tabular}{lcccccccc c}
\toprule
& \multicolumn{2}{c}{$a_0$} & \multicolumn{2}{c}{AUROC}
& \multicolumn{2}{c}{Review\%} & \multicolumn{3}{c}{Qualification under $u_{\mathrm{df}}$} \\
\cmidrule(lr){2-3}\cmidrule(lr){4-5}\cmidrule(lr){6-7}\cmidrule(lr){8-10}
\textbf{System} & $u$ & $u_{\mathrm{df}}$ & $u$ & $u_{\mathrm{df}}$
& $u$ & $u_{\mathrm{df}}$ & $\widehat R$ & 95\% LCB & Qual. \\
\midrule
\multicolumn{10}{@{}l}{\texttt{Claude Code}}\\
\quad Opus 5 & 75.7 & 69.6 & 0.680 & 0.691 & 22.4 & 37.9 & 0.842 & 0.813 & \checkmark \\
\quad Sonnet 5 & 72.5 & 65.1 & 0.681 & 0.687 & 29.6 & 41.6 & 0.841 & 0.812 & \checkmark \\
\addlinespace
\multicolumn{10}{@{}l}{\texttt{Codex}}\\
\quad GPT-5.6-Sol & 73.1 & 67.7 & 0.640 & 0.631 & 32.5 & 44.5 & 0.830 & 0.801 & \checkmark \\
\quad GPT-5.6-Luna & 69.9 & 64.3 & 0.645 & 0.631 & 41.6 & 56.0 & 0.829 & 0.802 & \checkmark \\
\quad GPT-5.5 & 75.7 & 70.9 & 0.652 & 0.664 & 21.3 & 35.7 & 0.852 & 0.823 & \checkmark \\
\quad GPT-5.4 & 72.8 & 68.3 & 0.601 & 0.630 & 39.2 & 46.9 & 0.825 & 0.796 & \checkmark \\
\quad GPT-5.4-mini & 67.5 & 57.3 & 0.628 & 0.624 & 49.1 & 67.2 & 0.855 & 0.832 & \checkmark \\
\addlinespace
\multicolumn{10}{@{}l}{\texttt{Gemini CLI}}\\
\quad Gemini-3.6-Flash & 71.7 & 58.7 & 0.619 & 0.612 & 38.4 & 100.0 & 0.900 & 0.900 & \checkmark \\
\quad Gemini-3.5-Flash & 70.4 & 58.7 & 0.618 & 0.597 & 38.7 & 100.0 & 0.900 & 0.900 & \checkmark \\
\quad Gemini-3.1-Pro & 70.9 & 58.7 & 0.563 & 0.571 & 100.0 & 100.0 & 0.900 & 0.900 & \checkmark \\
\addlinespace
\multicolumn{10}{@{}l}{\texttt{opencode}}\\
\quad DeepSeek-V4-Pro & 68.5 & 58.1 & 0.664 & 0.652 & 44.0 & 68.5 & 0.865 & 0.842 & \checkmark \\
\quad DeepSeek-V4-Flash & 68.8 & 56.3 & 0.602 & 0.633 & 37.1 & 84.5 & 0.891 & 0.876 & \checkmark \\
\quad GLM-5.2 & 75.7 & 66.1 & 0.652 & 0.678 & 10.9 & 44.8 & 0.811 & 0.781 & \checkmark \\
\quad Qwen-3.7-Plus & 66.4 & 54.1 & 0.711 & 0.679 & 61.9 & 61.9 & 0.815 & 0.789 & \checkmark \\
\quad MiniMax-M3 & 66.4 & 55.2 & 0.676 & 0.653 & 33.9 & 79.7 & 0.880 & 0.862 & \checkmark \\
\quad Kimi-K2.7-Code & 65.9 & 53.1 & 0.661 & 0.668 & 67.7 & 67.7 & 0.828 & 0.803 & \checkmark \\
\bottomrule
\end{tabular}
\caption{\textbf{Routing quality and held-out qualification under the
process-aware success definition.} $a_0$ and AUROC are on $S_{\mathrm{dev}}$
($n=375$); Review\%, $\widehat R$ and the LCB are on $S_{\mathrm{qual}}$ at
$Y=0.76$, $a_h=0.90$, with thresholds reselected at $Y+0.05$ under each
predicate. AUROC is computed from the threshold-based risk--coverage curve, so
values are directly comparable across the two columns.}
\label{tab:app-defectfree}
\end{table}

\subsection{Complete Per-System Results}
\label{app:clinicare-full-tables}

This section reports the complete versions of the main-text summaries:
per-system risk--coverage and qualification results for all evaluated
systems, and routing-score distributions with autonomous success broken down
by predicted and reference verdict class.

\paragraph{Qualification across targets.}
Table~\ref{tab:app-qualification-full} extends the single-target view of
Table~\ref{tab:clinicare-qualification} to three targets spanning the range in
which most systems have reachable deploying operating points.
The step-function behaviour noted in Section~\ref{sec:empirical-qualification-results}
is visible throughout: review burden is flat across adjacent targets for systems
whose confidence takes few distinct values, then jumps.
Opus 5 moves $0.8\to22.4\to37.9$ percent review across the three targets while
Gemini-3.1-Pro is pinned at $100\%$ for the upper two.

\paragraph{Routing behaviour by verdict class.}
Table~\ref{tab:app-class-conditional} breaks the qualification partition down by
predicted verdict.
Pooled across systems, autonomous success is strongly class-dependent
--- $0.856$ for \textit{Yes}, $0.657$ for \textit{No}, $0.544$ for
\textit{Indeterminate: Lack of Data} and $0.336$ for
\textit{Indeterminate: Medically Ambiguous} --- while mean stated confidence
varies far less ($0.894$, $0.891$, $0.839$, $0.797$).
Systems are therefore somewhat but not proportionately less confident on the
verdicts they get wrong more often.

The result that matters for the policy design of
Section~\ref{sec:empirical-policies} is that the routing signal retains ranking
power \emph{within} every predicted class: pooled within-class AUROC is $0.639$
for \textit{Yes}, $0.611$ for \textit{No}, $0.614$ for \textit{ILD} and $0.668$
for \textit{IMA}.
The signal is thus not a proxy for the predicted class.
A class-agnostic threshold rule is not discarding information that a
class-conditional rule would recover, which is the empirical basis for keeping
class semantics inside $\phi_w$ and out of the escalation policy.

\begin{table}[h!]
\centering
\small
\setlength{\tabcolsep}{4pt}
\begin{tabular}{lcccc cccc cccc}
\toprule
& \multicolumn{4}{c}{$Y=0.71$} & \multicolumn{4}{c}{$Y=0.76$} & \multicolumn{4}{c}{$Y=0.80$} \\
\cmidrule(lr){2-5}\cmidrule(lr){6-9}\cmidrule(lr){10-13}
\textbf{System}
& Rev\% & $\widehat R$ & LCB & Q
& Rev\% & $\widehat R$ & LCB & Q
& Rev\% & $\widehat R$ & LCB & Q \\
\midrule
\multicolumn{13}{@{}l}{\texttt{Claude Code}}\\
\quad Opus 5 & 0.8 & 0.759 & 0.720 & \checkmark & 22.4 & 0.839 & 0.807 & \checkmark & 37.9 & 0.871 & 0.844 & \checkmark \\
\quad Sonnet 5 & 12.0 & 0.817 & 0.783 & \checkmark & 29.6 & 0.856 & 0.826 & \checkmark & 41.6 & 0.881 & 0.856 & \checkmark \\
\addlinespace
\multicolumn{13}{@{}l}{\texttt{Codex}}\\
\quad GPT-5.6-Sol & 22.1 & 0.802 & 0.768 & \checkmark & 32.5 & 0.842 & 0.812 & \checkmark & 44.5 & 0.859 & 0.832 & \checkmark \\
\quad GPT-5.6-Luna & 28.3 & 0.748 & 0.712 & \checkmark & 41.6 & 0.796 & 0.764 & \checkmark & 56.0 & 0.853 & 0.828 & \checkmark \\
\quad GPT-5.5 & 1.9 & 0.774 & 0.736 & \checkmark & 21.3 & 0.843 & 0.811 & \checkmark & 38.1 & 0.877 & 0.850 & \checkmark \\
\quad GPT-5.4 & 12.5 & 0.731 & 0.693 & -- & 39.2 & 0.827 & 0.797 & \checkmark & 59.2 & 0.861 & 0.836 & \checkmark \\
\quad GPT-5.4-mini & 29.1 & 0.800 & 0.767 & \checkmark & 49.1 & 0.850 & 0.822 & \checkmark & 67.2 & 0.879 & 0.858 & \checkmark \\
\addlinespace
\multicolumn{13}{@{}l}{\texttt{Gemini CLI}}\\
\quad Gemini-3.6-Flash & 38.4 & 0.850 & 0.821 & \checkmark & 38.4 & 0.850 & 0.821 & \checkmark & 100.0 & 0.900 & 0.900 & \checkmark \\
\quad Gemini-3.5-Flash & 38.7 & 0.847 & 0.818 & \checkmark & 38.7 & 0.847 & 0.818 & \checkmark & 100.0 & 0.900 & 0.900 & \checkmark \\
\quad Gemini-3.1-Pro & 33.6 & 0.838 & 0.808 & \checkmark & 100.0 & 0.900 & 0.900 & \checkmark & 100.0 & 0.900 & 0.900 & \checkmark \\
\addlinespace
\multicolumn{13}{@{}l}{\texttt{opencode}}\\
\quad DeepSeek-V4-Pro & 44.0 & 0.839 & 0.810 & \checkmark & 44.0 & 0.839 & 0.810 & \checkmark & 68.5 & 0.883 & 0.863 & \checkmark \\
\quad DeepSeek-V4-Flash & 37.1 & 0.832 & 0.802 & \checkmark & 37.1 & 0.832 & 0.802 & \checkmark & 84.5 & 0.891 & 0.876 & \checkmark \\
\quad GLM-5.2 & 1.1 & 0.716 & 0.676 & -- & 10.9 & 0.762 & 0.725 & -- & 37.6 & 0.845 & 0.816 & \checkmark \\
\quad Qwen-3.7-Plus & 22.7 & 0.775 & 0.739 & \checkmark & 61.9 & 0.874 & 0.851 & \checkmark & 61.9 & 0.874 & 0.851 & \checkmark \\
\quad MiniMax-M3 & 33.9 & 0.809 & 0.777 & \checkmark & 33.9 & 0.809 & 0.777 & \checkmark & 61.1 & 0.872 & 0.849 & \checkmark \\
\quad Kimi-K2.7-Code & 29.1 & 0.806 & 0.773 & \checkmark & 67.7 & 0.860 & 0.837 & \checkmark & 67.7 & 0.860 & 0.837 & \checkmark \\
\bottomrule
\end{tabular}
\caption{\textbf{Held-out qualification across three reliability targets.}
$N_{\mathrm{qual}}=375$, $a_h=0.90$, thresholds selected on $S_{\mathrm{dev}}$
at $Y+0.05$ and frozen. Rev\% is the fraction of qualification cases routed to
human review, LCB the one-sided $95\%$ bound of
Appendix~\ref{app:clinicare-lcb-procedure}, and Q whether the policy qualifies.
The $Y=0.76$ block reproduces Table~\ref{tab:clinicare-qualification}; the
break-even assumption $a_h^{\min}$ is omitted here for width and is reported at
$Y=0.76$ in that table. Every policy that fails to qualify does so under no
$a_h\in[0,1]$.}
\label{tab:app-qualification-full}
\end{table}

\begin{table}[h!]
\centering
\small
\begin{tabular}{lccccccccc}
\toprule
& \multicolumn{2}{c}{\textit{Yes}} & \multicolumn{2}{c}{\textit{No}}
& \multicolumn{2}{c}{\textit{ILD}} & \multicolumn{2}{c}{\textit{IMA}} & \\
\cmidrule(lr){2-3}\cmidrule(lr){4-5}\cmidrule(lr){6-7}\cmidrule(lr){8-9}
\textbf{System} & \% & Succ. & \% & Succ. & \% & Succ. & \% & Succ.
& $|\{s_i\}|$ \\
\midrule
\multicolumn{10}{@{}l}{\texttt{Claude Code}}\\
\quad Opus 5 & 42.4 & 86.8 & 38.4 & 71.5 & 16.0 & 58.3 & 3.2 & 58.3 & 25 \\
\quad Sonnet 5 & 43.7 & 85.4 & 36.5 & 73.0 & 16.0 & 63.3 & 3.7 & 28.6 & 26 \\
\addlinespace
\multicolumn{10}{@{}l}{\texttt{Codex}}\\
\quad GPT-5.6-Sol & 38.4 & 91.7 & 39.2 & 67.3 & 18.7 & 54.3 & 3.7 & 42.9 & 19 \\
\quad GPT-5.6-Luna & 34.1 & 86.7 & 41.1 & 61.0 & 18.4 & 40.6 & 6.4 & 25.0 & 21 \\
\quad GPT-5.5 & 41.1 & 90.9 & 42.1 & 68.4 & 13.9 & 61.5 & 2.9 & 63.6 & 25 \\
\quad GPT-5.4 & 35.7 & 88.8 & 42.1 & 65.8 & 17.9 & 47.8 & 4.3 & 31.2 & 32 \\
\quad GPT-5.4-mini & 39.5 & 83.8 & 42.7 & 60.0 & 12.3 & 45.7 & 5.6 & 19.0 & 33 \\
\addlinespace
\multicolumn{10}{@{}l}{\texttt{Gemini CLI}}\\
\quad Gemini-3.6-Flash & 42.9 & 86.3 & 39.7 & 67.1 & 13.3 & 66.0 & 4.0 & 40.0 & 5 \\
\quad Gemini-3.5-Flash & 44.3 & 84.9 & 40.5 & 66.4 & 11.7 & 63.6 & 3.5 & 38.5 & 5 \\
\quad Gemini-3.1-Pro & 39.7 & 87.2 & 37.1 & 71.2 & 21.3 & 56.2 & 1.9 & 71.4 & 4 \\
\addlinespace
\multicolumn{10}{@{}l}{\texttt{opencode}}\\
\quad DeepSeek-V4-Pro & 38.7 & 86.2 & 41.6 & 62.2 & 14.7 & 54.5 & 4.8 & 33.3 & 15 \\
\quad DeepSeek-V4-Flash & 49.1 & 81.0 & 35.7 & 70.1 & 11.7 & 56.8 & 3.5 & 30.8 & 10 \\
\quad GLM-5.2 & 45.9 & 80.8 & 39.2 & 68.0 & 11.2 & 57.1 & 3.7 & 35.7 & 23 \\
\quad Qwen-3.7-Plus & 35.2 & 87.9 & 42.4 & 58.5 & 14.9 & 39.3 & 7.5 & 35.7 & 18 \\
\quad MiniMax-M3 & 40.0 & 79.3 & 41.9 & 62.4 & 9.9 & 56.8 & 8.3 & 22.6 & 19 \\
\quad Kimi-K2.7-Code & 39.7 & 84.6 & 43.5 & 61.3 & 9.1 & 55.9 & 7.7 & 24.1 & 14 \\
\bottomrule
\end{tabular}
\caption{\textbf{Predicted-class mix and within-class autonomous success on the
qualification partition.} \% is the share of the $375$ cases receiving each
predicted verdict; Succ.\ is four-way verdict accuracy within that share.
\textit{ILD} and \textit{IMA} abbreviate the two indeterminate classes of
Equation~\ref{eq:clinicare-fourway-labels}. $|\{s_i\}|$ is the number of
distinct stated-confidence values the system emits on this partition, the
granularity limit discussed in Section~\ref{sec:empirical-routing}.}
\label{tab:app-class-conditional}
\end{table}

\subsection{Cost-of-Conflation Control: Always-Review-Indeterminate}
\label{app:conflation-control}

The main-text oversight policy acts only on the routing signal and is agnostic
to the four adjudication classes (Section~\ref{sec:empirical-policies}).
As a negative control, we contrast it with a policy that instead escalates on
the \emph{predicted class}: every indeterminate adjudication is routed to human
review regardless of confidence,

\begin{equation}
\pi_{\mathrm{IR}}(s_i,\widehat y_i)
=
\begin{cases}
\mathrm{review},
&
\widehat y_i
\in
\{
\mathrm{Indeterminate{:}\ Lack\ of\ Data},
\mathrm{Indeterminate{:}\ Medically\ Ambiguous}
\},
\\[3pt]
\mathrm{accept},
&
\widehat y_i\in\{\mathrm{Yes},\mathrm{No}\}
\ \land\ s_i\geq\tau,
\\[3pt]
\mathrm{review},
&
\text{otherwise}.
\end{cases}
\label{eq:clinicare-review-indeterminate-policy}
\end{equation}

This baseline operationalizes the conflation \EFE{} is designed to avoid: it
treats a task-level abstention, a possibly-correct indeterminate verdict, an
\emph{outcome} scored by $\phi_w$, as if it were a deployment-level escalation,
a \emph{decision} that should depend on the routing signal.
We select and freeze $\tau$ for $\pi_{\mathrm{IR}}$ exactly as for the
signal-only policy, and report the comparison in
Table~\ref{tab:app-conflation}.

\paragraph{The control is not more expensive on this benchmark.}
We had expected the conflation to show up as wasted oversight, and it does not.
Averaged over the $496$ policies in the sweep, $\pi_{\mathrm{IR}}$ routes
$48.9\%$ of cases to review against $53.0\%$ for the signal-only policy, and it
requires strictly more review in only $206$ of $496$ cells.
The single-target comparison could be dismissed as threshold-grid overshoot, so
the right-hand columns of Table~\ref{tab:app-conflation} instead report, for
each policy, the minimum review burden at which it \emph{achieves} a given
reliability on the qualification partition, which removes the grid effect.
The control is still not dominated: it needs more review on $10$ of $16$ systems
at $\widehat R=0.80$ and $6$ of $15$ at $\widehat R=0.85$, with mean differences
of $-2.2$ and $-5.2$ percentage points.

The reason is visible in Table~\ref{tab:app-class-conditional}.
Indeterminate adjudications really are much less often correct than definite
ones on this benchmark ($0.493$ against $0.757$ pooled on the qualification
partition), so predicted class carries genuine information about autonomous
failure.
Escalating on it is not waste, and we report that rather than presenting a
control the data does not support.

\paragraph{What the control does cost.}
The objection to $\pi_{\mathrm{IR}}$ is structural rather than budgetary.
Because the class constraint is fixed, the policy exposes one hard floor on
review burden per system --- no threshold can accept an indeterminate case ---
so it does not trace a frontier over the reliability range the way
Equation~\ref{eq:clinicare-fourway-global-policy} does.
It also inherits the benchmark's class-success profile as an assumption: it is
competitive here only because indeterminate verdicts happen to be unreliable in
this cohort, a property of these labels rather than of the deployment, and one
that no part of the policy measures or would detect if it changed.
The signal-based policy makes the same trade explicit and tunable, and
Appendix~\ref{app:clinicare-full-tables} shows the routing signal remains
informative \emph{within} each class, so the two are complementary sources of
information rather than substitutes.
A deployment free to use both would do better than either; the claim defended
here is only that the escalation decision should be driven by a measured
routing signal rather than inferred from an outcome class.

\begin{table}[h!]
\centering
\small
\begin{tabular}{lcccc cccc}
\toprule
& \multicolumn{4}{c}{At $Y=0.76$} & \multicolumn{4}{c}{Min.\ Review\% to achieve $\widehat R$} \\
\cmidrule(lr){2-5}\cmidrule(lr){6-9}
& \multicolumn{2}{c}{Review\%} & & & \multicolumn{2}{c}{$\widehat R=0.80$}
& \multicolumn{2}{c}{$\widehat R=0.85$} \\
\cmidrule(lr){2-3}\cmidrule(lr){6-7}\cmidrule(lr){8-9}
\textbf{System} & $\pi_\tau$ & $\pi_{\mathrm{IR}}$ & $\widehat R_{\mathrm{IR}}$
& Qual. & $\pi_\tau$ & $\pi_{\mathrm{IR}}$ & $\pi_\tau$ & $\pi_{\mathrm{IR}}$ \\
\midrule
\multicolumn{9}{@{}l}{\texttt{Claude Code}}\\
\quad Opus 5 & 22.4 & 19.2 & 0.815 & \checkmark & 10.4 & 19.2 & 37.9 & 34.4 \\
\quad Sonnet 5 & 29.6 & 21.1 & 0.827 & \checkmark & 12.0 & 19.7 & 29.6 & 34.7 \\
\addlinespace
\multicolumn{9}{@{}l}{\texttt{Codex}}\\
\quad GPT-5.6-Sol & 32.5 & 26.1 & 0.827 & \checkmark & 22.1 & 22.4 & 44.5 & 41.1 \\
\quad GPT-5.6-Luna & 41.6 & 28.5 & 0.790 & -- & 56.0 & 33.6 & 56.0 & 54.9 \\
\quad GPT-5.5 & 21.3 & 17.6 & 0.814 & \checkmark & 15.5 & 16.8 & 35.7 & 28.0 \\
\quad GPT-5.4 & 39.2 & 31.2 & 0.819 & \checkmark & 29.3 & 23.2 & 59.2 & 44.3 \\
\quad GPT-5.4-mini & 49.1 & 45.9 & 0.853 & \checkmark & 29.1 & 31.2 & 53.6 & 45.9 \\
\addlinespace
\multicolumn{9}{@{}l}{\texttt{Gemini CLI}}\\
\quad Gemini-3.6-Flash & 38.4 & 45.6 & 0.864 & \checkmark & 38.4 & 45.6 & 43.5 & 45.6 \\
\quad Gemini-3.5-Flash & 38.7 & 45.3 & 0.859 & \checkmark & 38.7 & 45.3 & 43.2 & 45.3 \\
\quad Gemini-3.1-Pro & 100.0 & 52.8 & 0.897 & \checkmark & 33.6 & 23.2 & \textemdash & 52.8 \\
\addlinespace
\multicolumn{9}{@{}l}{\texttt{opencode}}\\
\quad DeepSeek-V4-Pro & 44.0 & 54.9 & 0.860 & \checkmark & 44.0 & 54.9 & 68.5 & 54.9 \\
\quad DeepSeek-V4-Flash & 37.1 & 45.3 & 0.856 & \checkmark & 37.1 & 26.1 & 84.5 & 45.3 \\
\quad GLM-5.2 & 10.9 & 17.1 & 0.775 & -- & 32.0 & 23.5 & 44.8 & 40.3 \\
\quad Qwen-3.7-Plus & 61.9 & 38.9 & 0.820 & \checkmark & 61.9 & 30.4 & 61.9 & 64.0 \\
\quad MiniMax-M3 & 33.9 & 39.2 & 0.825 & \checkmark & 33.9 & 39.2 & 60.8 & 64.8 \\
\quad Kimi-K2.7-Code & 67.7 & 37.9 & 0.818 & \checkmark & 29.1 & 33.6 & 67.7 & 70.4 \\
\bottomrule
\end{tabular}
\caption{\textbf{Signal-based routing against the always-review-indeterminate
control}, $a_h=0.90$, $N_{\mathrm{qual}}=375$. Left block: both policies
selected on $S_{\mathrm{dev}}$ at $Y+0.05$ and frozen, evaluated at $Y=0.76$.
Right block: the minimum review burden at which each policy attains the stated
reliability among its own reachable operating points on $S_{\mathrm{qual}}$,
which removes the threshold-grid overshoot that confounds the single-target
comparison; \textemdash{} marks a reliability no operating point of that policy
reaches.}
\label{tab:app-conflation}
\end{table}